\documentclass[10pt,twocolumn,letterpaper]{article}

\usepackage[pagenumbers]{cvpr} 

\definecolor{cvprblue}{rgb}{0.21,0.49,0.74}
\usepackage[pagebackref,breaklinks,colorlinks,allcolors=cvprblue]{hyperref}
\usepackage{booktabs}
\usepackage{graphicx}
\usepackage{multirow}
\usepackage{bm}
\usepackage{adjustbox}
\usepackage{colortbl}

\usepackage{array,booktabs,tabularx,multirow}
\newcolumntype{Y}{>{\centering\arraybackslash}X}
\newcommand{\metrichead}[1]{%
  \begin{tabular}[c]{@{}c@{}}#1\end{tabular}%
}

\definecolor{bestpurple}{HTML}{C7C4FA}
\definecolor{secondpurple}{HTML}{E9E7FC}
\newcommand{\best}[1]{\cellcolor{bestpurple}\textbf{#1}}
\newcommand{\second}[1]{\cellcolor{secondpurple}#1}
\newcommand{\methodname}{DriveRender}

\def\confName{CVPR}
\def\confYear{2027}

\title{DriveRender: Geometry-Grounded Autoregressive \\ Generative Rendering for Driving Scenes}

\author{%
  {\large
    Jiayi Xu$^{1,2,*}$, Zekai Gu$^{1,*}$, Yingshuang Zou$^{1}$,
    Jiahao Lu$^{1}$, Li Ziang$^{1}$, Tianyu Liu$^{1,2}$%
  }\\[0.2em]
  {\large
    Haiping Wang$^{1}$, Wei Wu$^{2}$, Sai-Kit Yeung$^{1}$,
    Maosheng Ye$^{2}$, Tongyi Cao$^{2}$, Yuan Liu$^{1,\ddagger}$%
  }\\[0.55em]
  {\normalsize
    $^{1}$The Hong Kong University of Science and Technology
    \qquad $^{2}$DeepRoute%
  }\\[0.35em]
  {\small
    $^{*}$Equal contribution \qquad $^{\ddagger}$Corresponding author%
  }%
}

\makeatletter
\g@addto@macro\@maketitle{%
    \par
    \vspace{-25pt}
    \noindent
    \makebox[\textwidth][c]{%
        \begin{minipage}{0.88\textwidth}
            \centering
            \includegraphics[width=\linewidth]{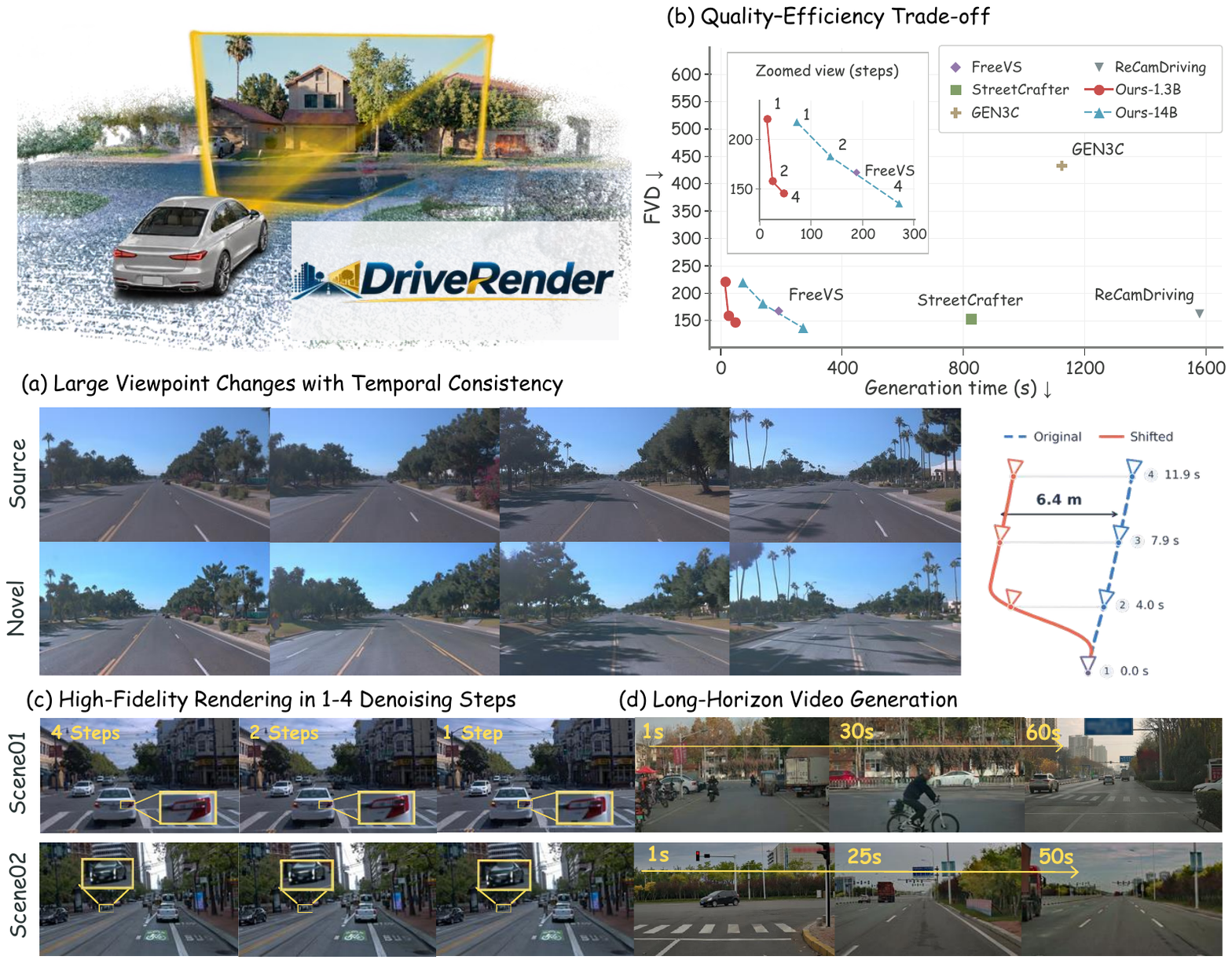}
            \captionsetup{type=figure,skip=-10pt}
            \caption{\textbf{Overview of DriveRender}. DriveRender synthesizes novel-view driving videos along user-specified camera trajectories, featuring
(a) temporal consistency under large viewpoint changes,
(b) flexible quality--efficiency trade-offs,
(c) high-fidelity rendering in 1--4 denoising steps and
(d) long-horizon autoregressive generation for up to 60 seconds.}
            \label{fig:teaser}
        \end{minipage}%
    }
    \par
    \vspace{4pt}
}
\makeatother

\begin{document}
\maketitle
\vspace*{2pt}
\begin{abstract}
Novel-view synthesis (NVS) empowers realistic simulation along alternative camera trajectories in autonomous driving. However, 3DGS approaches continue to struggle with per-scene parameter estimation and restricted visibility under large viewpoint variations. While video diffusion priors offer a promising alternative, the iterative denoising incurs substantial computational costs, and autoregressive generation can lead to accumulating appearance drift over long sequences. In this paper, we present \textbf{DriveRender}, a geometry-grounded generative rendering framework for efficient and continuous driving NVS. Our core insight is that explicit geometric conditioning from point-cloud projections strictly constrains scene structure, allowing aggressive denoising step reduction without sacrificing fidelity. Built upon a coverage-aware warping strategy and a staged distillation pipeline, our framework preserves long-horizon temporal coherence with minimal inference steps. Extensive experiments on diverse driving datasets demonstrate that DriveRender achieves high-quality, temporally consistent novel-trajectory synthesis in only 1--4 denoising steps, unlocking efficient data generation for autonomous driving simulation. Project page: \url{https://xujiayi777.github.io/DriveRender/}
\end{abstract}

    
\section{Introduction}
Novel-view synthesis (NVS) from recorded driving sequences enables the exploration of real-world driving environments along alternative camera trajectories, providing valuable capabilities for autonomous driving simulation~\cite{wang2025freevs,yan2025streetcrafter,zhou2024drivinggaussian, yan2024streetgaussian, fan2025freesim,yang2025drivex,sun2024drivescenegen}. Given a recorded drive, the goal is to synthesize observations from novel viewpoints while preserving the geometry, appearance, and dynamics of the original scene. This allows recorded driving data to be reused across diverse ego-motion scenarios without repeated data collection~\cite{tian2026simscale}.

Traditional driving NVS methods mainly rely on scene reconstruction~\cite{wang2024dust3r,zhang2025monst3r,lu2025align3r,wang2025vggt,wang2026pi,lin2025depth}, particularly NeRF~\cite{mildenhall2021nerf} and 3DGS~\cite{kerbl20233d}, but often suffer from degraded rendering quality under large viewpoint changes due to insufficient multi-view observations~\cite{ yang2024emernerf,zhou2024drivinggaussian, yan2024streetgaussian}.
To deal with this limitation, recent approaches integrate generative priors into reconstruction pipelines to recover plausible content in under-observed regions~\cite{fan2025freesim,yang2025drivex,lu2026gae}.
Beyond refining explicit reconstructions, a promising direction is to directly employ video diffusion models as geometry-conditioned renderers, leveraging pretrained generative priors to synthesize plausible appearances
in unseen regions~\cite{ren2025gen3c, lin2026vista4d,yu2025trajectorycrafter}. These works demonstrated strong performance on the short-term rendering of limited-sized scenes, but how to generalize such generative rendering techniques to long-term and dynamic driving scenes still requires additional exploration.   

Specifically, applying generative models to driving NVS presents three key challenges. (1) \textbf{Scene fidelity}. Despite their strong generative priors, video generation models may hallucinate content or introduce geometric inconsistencies, failing to faithfully preserve the recorded scene under novel camera trajectories~\cite{ren2025gen3c,wu2025difix3d+,yan2025streetcrafter}. (2) \textbf{Long-horizon consistency}. Existing generative NVS methods often operate within short temporal windows~\cite{yan2025streetcrafter,li2025recamdriving}, while extending generation autoregressively can accumulate errors and cause appearance and structural drift over time~\cite{zhang2025epona,zhang2026horizondrive}. (3) \textbf{Rendering efficiency}. High-quality diffusion-based rendering typically requires many iterative denoising steps, resulting in substantial inference costs~\cite{yan2025streetcrafter,ren2025gen3c,li2025recamdriving} that become increasingly prohibitive for long driving sequences. Addressing these challenges requires a generative renderer that is scene-faithful, temporally consistent over extended trajectories, and computationally efficient.

Motivated by these challenges, we propose \textbf{\methodname{}}, a geometry-grounded autoregressive generative rendering framework for efficient and continuous driving NVS. Built upon a pretrained video diffusion model, \methodname{} leverages geometric and appearance cues from the recorded driving video to guide novel-view synthesis, while generating video chunks autoregressively to support continuous navigation along new camera trajectories.

To ensure \textbf{scene fidelity}, we design complementary geometric and reference video conditions that anchor generation to the recorded environment. While point-cloud projections establish pixel-level spatial correspondences, directly aggregating multiple point clouds introduces geometric conflicts~\cite{yan2025streetcrafter}, whereas relying on a single frame leaves large uncovered regions~\cite{xi2026geov2v, ren2025gen3c, shen2026lyra}.
We therefore adopt a \textit{coverage-aware warping strategy} that selectively fills missing regions using additional source frames, improving spatial coverage while reducing artifacts from inconsistent observations. Moreover, geometric projections can be incomplete or fragmented due to occlusions and reconstruction errors, losing fine appearance and semantic details. We therefore complement them with reference video frames and encode their viewing geometry through Pl\"ucker ray embeddings in a shared coordinate system with the target views. Together, these conditions provide complementary spatial guidance and appearance cues for faithful scene rendering.

To achieve \textbf{efficient and long-horizon rendering}, we develop a staged distillation strategy tailored to geometry-conditioned generation. We first combine joint text--geometry guidance distillation with progressive step reduction to preserve visual details
with only 1--4 denoising steps. For long-horizon consistency, we further introduce \textit{geometry-anchored rollout distillation}. Unlike general autoregressive video generation~\cite{huang2026self,cui2026selfplus,yang2025longlive,zhu2026causal,zhao2026causalplus}, where scene consistency relies heavily on generated histories, our renderer continuously receives geometric and appearance anchors from the recorded scene. Exploiting this property, we distill short consecutive segments sampled from extended autoregressive rollouts, using a bidirectional teacher to provide joint supervision under the same scene conditions. This enables efficient rollout adaptation without requiring the teacher to process the entire generation history, while mitigating accumulated appearance drift.

Extensive experiments on driving datasets demonstrate that \methodname{} achieves high-quality novel-trajectory synthesis with improved scene consistency and long-horizon temporal coherence while substantially reducing diffusion sampling costs. Our results show that a geometry-grounded diffusion model can be efficient for recorded driving environments. With only 1 to 4 denoising steps, \methodname{} reduces the computational cost of novel-trajectory video synthesis, enabling more efficient data generation for driving simulation.

\section{Related Work}
\subsection{Driving Novel View Synthesis}
Driving scene reconstruction methods based on neural fields~\cite{yang2023unisim,yang2024emernerf,tonderski2024neurad} and Gaussian representations~\cite{zhou2024drivinggaussian,yan2024streetgaussian,chen2025omnire} enable realistic rendering but often struggle with poorly observed regions beyond recorded trajectories. To improve view extrapolation, several approaches incorporate generative priors into scene optimization~\cite{hwang2024vegs,zhao2025drivedreamer4d,ni2025recondreamer,yang2025drivex,fan2025freesim,zou2025mudg}. A complementary direction directly conditions video generation on geometric evidence~\cite{li2025uniscene}. FreeVS~\cite{wang2025freevs} and StreetCrafter~\cite{yan2025streetcrafter} leverage projected LiDAR observations for novel-view generation, while ReCamDriving~\cite{li2025recamdriving} employs 3DGS renderings and cross-trajectory supervision. GEN3C~\cite{ren2025gen3c} introduces a point-cloud cache to provide persistent geometric guidance, and GeoV2V~\cite{xi2026geov2v} combines LiDAR and depth-derived geometry with source-video conditioning. Despite their advances in viewpoint control and visual realism, these methods have yet to fully address the combined demands of efficient rendering and consistent scene appearance over extended camera trajectories.

\subsection{Autoregressive Video Diffusion and Distillation}
Diffusion acceleration has been explored through progressive distillation~\cite{salimans2022progressive}, guidance distillation~\cite{meng2023distillation}, consistency models~\cite{song2023consistency}, and distribution matching distillation~\cite{yin2024one,yin2024improved}. Recent methods extend these techniques to autoregressive video generation, including teacher--student distillation~\cite{yin2025slow}, on-policy training~\cite{huang2026self}, and few-step distillation~\cite{zhu2026causal,zhao2026causalplus}. Long-horizon generation is further explored through attention-based memory mechanisms~\cite{liu2026rolling,yang2025longlive} and improved autoregressive training~\cite{cui2026selfplus}. Helios~\cite{yuan2026helios} combines history-conditioned autoregressive generation with context compression and few-step distillation to enable real-time long-video synthesis. However, unlike general video generation, geometry-conditioned rendering benefits from strong spatial constraints that reduce the burden of scene synthesis, creating opportunities for more aggressive sampling reduction. Directly applying existing distillation strategies to this setting still blurs fine details and introduces appearance drift over extended rollouts. Motivated by these challenges, \methodname{} adopts a staged distillation strategy that first preserves scene details under reduced sampling budgets and then adapts the renderer to self-generated histories through geometry-anchored rollout distillation, promoting temporal consistency across generation windows.

\section{Method}

\subsection{Overview}
\label{subsec:overview}

\begin{figure*}[t]
    \centering
    \includegraphics[width=\textwidth]{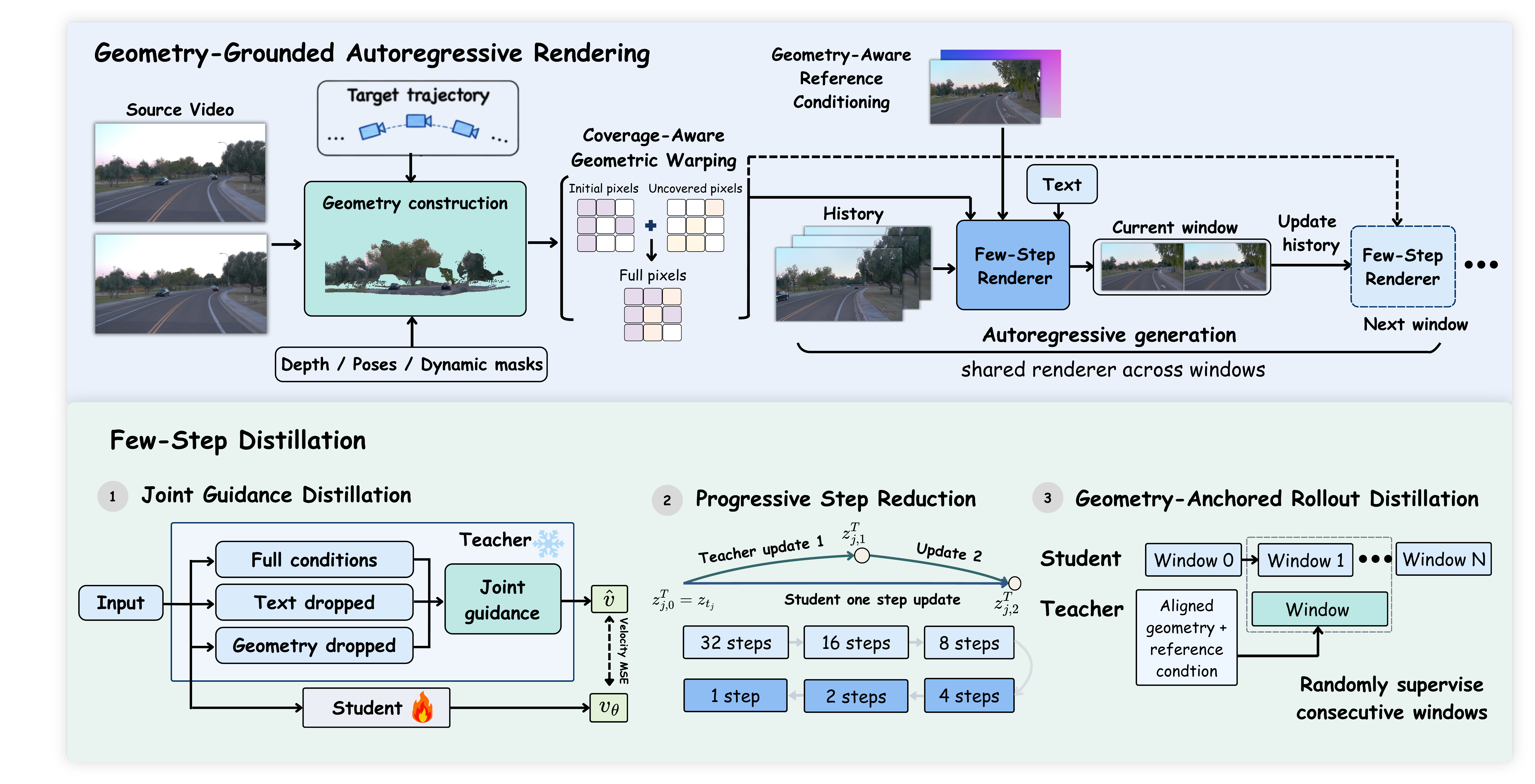}                     
    \caption{
    Overview of \methodname{}.
    \textbf{Top}: Geometry-conditioned autoregressive rendering from a recorded driving video, using projected geometry, reference images, and generated history to synthesize consecutive video windows along a target camera trajectory.
    \textbf{Bottom}: Three-stage distillation comprising joint guidance distillation, progressive step reduction for few-step sampling, and geometry-anchored rollout distillation.
    }
    \label{fig:framework}
    \vspace{-10pt}
\end{figure*}

\noindent\textbf{Task Formulation.}
Given a recorded driving video $\mathcal{I}=\{\mathbf{I}_i\}_{i=1}^{N}$, our goal is to enable virtual navigation through the recorded environment along a user-specified camera trajectory $\mathcal{P}^{\mathrm{tar}}= \{\mathbf{P}^{\mathrm{tar}}_t\}_{t=1}^{T}$.
Specifically, we aim to synthesize a novel-view video $\widehat{\mathcal{V}}$ that follows the target trajectory while preserving the scene geometry, appearance, and dynamics observed in the source video:
\begin{equation}
    \widehat{\mathcal{V}} = \mathcal{R}_{\theta}(\mathcal{I}, \mathcal{P}^{\mathrm{tar}}, c_{\mathrm{text}}),
\end{equation}
where $\mathcal{R}_{\theta}$ denotes our generative renderer and $c_{\mathrm{text}}$ is a global text condition. We first employ Pi3X~\cite{wang2026pi} to estimate colored 3D point clouds from the source video, providing an initial geometric scaffold for novel-view rendering.

\noindent\textbf{Framework Overview.}
As illustrated in~\cref{fig:framework}, we propose \methodname{}, a geometry-conditioned autoregressive video generation framework built upon a pretrained video diffusion transformer~\cite{peebles2023scalable,wan2025wan}. At each generation window, the model synthesizes a video chunk conditioned on three complementary sources of information: (1) pixel-aligned geometric projections that establish spatial correspondences with the recorded scene, (2) reference video frames that provide appearance and semantic context, and (3) previously generated frames that maintain temporal continuity. The generated chunk is then incorporated into the history for subsequent autoregressive rendering, enabling continuous synthesis along the target trajectory. 

In the following, we introduce how we construct the 3D geometric condition (Sec.~\ref{subsec:geometry_conditioning}).
Then, to achieve efficient generation, we apply joint guidance distillation and progressive step reduction to obtain a few-step renderer (Sec.~\ref{subsec:few-step_distillation}). 
Finally, we perform autoregressive rollout distillation to mitigate accumulated errors and appearance drift during long-horizon generation (Sec.~\ref{subsec:ar_distillation}).

\subsection{Geometric Condition Construction}
\label{subsec:geometry_conditioning}

To faithfully render the recorded scene from novel viewpoints, we construct pixel-aligned geometric conditions that establish spatial correspondences between source observations and target views. We further incorporate reference video frames to provide complementary appearance and semantic information, improving scene consistency when geometric projections are incomplete.

\noindent\textbf{Coverage-Aware Geometric Warping.}
A straightforward approach is to aggregate point clouds from multiple source frames and project them into the target view. However, imperfect geometric reconstruction and alignment across frames can introduce conflicting observations, leading to ghosting and degraded rendering quality. Alternatively, projecting only the temporally closest source frame avoids such conflicts but often leaves large unobserved regions, particularly when the target camera trajectory deviates significantly from the recorded trajectory.

\begin{figure}[t]
    \centering
    \includegraphics[width=\columnwidth]{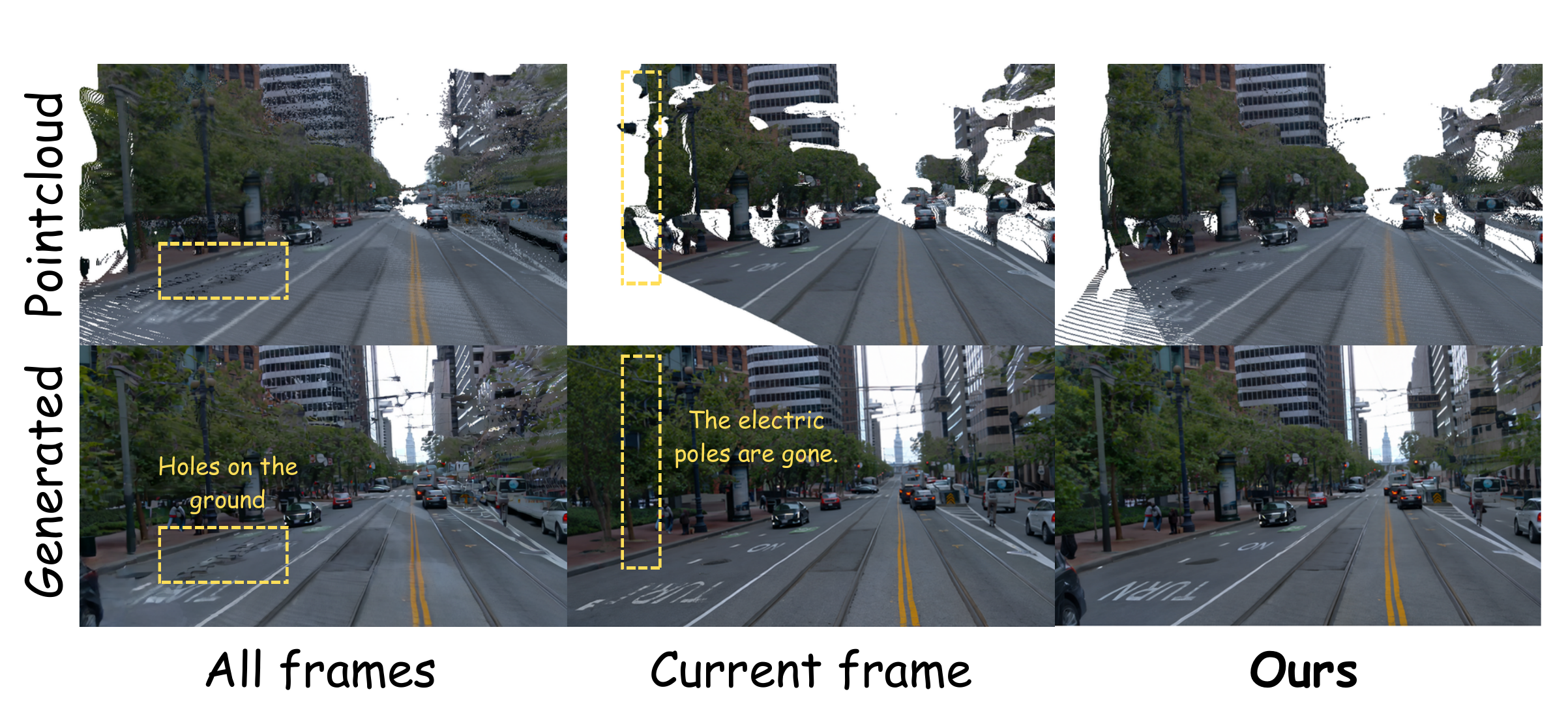}               
    \caption{Qualitative comparisons with different warping strategies.} 
    \label{fig:waring_strategy}
    \vspace{-15pt}
\end{figure}

To balance spatial coverage and geometric consistency, we adopt a coverage-aware warping strategy. We first project the point cloud from the temporally closest source frame into the target view, obtaining an initial warped image $\mathbf{G}_t$ and validity mask $\mathbf{M}_t$.
We then progressively incorporate additional source frames, filling only previously uncovered pixels:
\begin{equation}
\mathbf{G}_t \leftarrow
\mathbf{M}_t \odot \mathbf{G}_t
+ (1-\mathbf{M}_t)\odot
\mathbf{W}_{i\rightarrow t},
\label{eq:coverage_aware_warping}
\end{equation}
where $\mathbf{W}_{i\rightarrow t}$ denotes the projection from source frame $i$, and the validity mask $\mathbf{M}_t$ is updated after each projection. This selective aggregation improves geometric coverage while reducing artifacts from inconsistent multi-frame projections. To avoid ghosting from moving objects, we retain dynamic points only from the temporally corresponding frame and use static points from other frames as shown in~\cref{fig:waring_strategy}.

\noindent\textbf{Geometry-Aware Reference Conditioning.} Although geometric warping provides pixel-level spatial correspondences, its projections can be incomplete or fragmented due to occlusions and reconstruction errors, often losing important object appearance and semantic details. We therefore additionally condition the renderer on reference frames selected from the recorded video, which provide more complete visual evidence of the scene.

However, reference frames are captured from different viewpoints and are not directly aligned with the target video. To provide viewpoint-aware conditioning, we associate each reference token
with a Pl\"ucker ray embedding computed from its camera intrinsics and pose. These rays are expressed in the same coordinate system as the target-view rays, providing geometric cues about the viewing direction and camera location of each observation. The ray embeddings are projected into the hidden feature space and added to the token features before self-attention in each
transformer block.


\subsection{Few-Step Distillation}
\label{subsec:few-step_distillation}

Although explicit geometric conditioning reduces the uncertainty of novel-view synthesis, high-quality video diffusion models still require multiple denoising steps and guidance evaluations, resulting in high rendering costs. We therefore adopt a two-stage distillation strategy to accelerate rendering while preserving visual fidelity.

\noindent\textbf{Joint Text and Geometry Guidance Distillation.}
Geometric guidance encourages consistency with the recorded scene, while text guidance leverages the pretrained generative prior for plausible completion. Since geometric projections may contain holes and reconstruction errors, we separately control the strengths of these two guidance signals and distill their combined prediction into a single student model. Specifically, the guided teacher prediction is
\begin{equation}
\begin{aligned}
\widehat{\mathbf{v}}_{\psi}
&= \mathbf{v}_{\mathrm{base}}
+ w_{\mathrm{text}}
\left[
\mathbf{v}_{\psi}(c_{\mathrm{text}},c_g)
-\mathbf{v}_{\psi}(\emptyset,c_g)
\right]\\[-2pt]
&\quad
+ w_g
\left[
\mathbf{v}_{\psi}(c_{\mathrm{text}},c_g)
-\mathbf{v}_{\psi}(c_{\mathrm{text}},\emptyset)
\right],
\end{aligned}
\label{eq:joint_text_geometry_guidance}
\end{equation}
where $c_{\mathrm{text}}$ and $c_g$ denote the text and geometry-related conditions, respectively, and $\mathbf{v}_{\psi}$ is the frozen teacher. We define $\mathbf{v}_{\mathrm{base}}=\alpha\mathbf{v}_{\psi}(\emptyset,c_g)+(1-\alpha)\mathbf{v}_{\psi}(c_{\mathrm{text}},\emptyset)$, where $\alpha=w_{\mathrm{text}}/(w_{\mathrm{text}}+w_g)$. The student matches the guided teacher prediction,
reducing three network evaluations to one per step.

\noindent\textbf{Progressive Step Reduction (PSR).}
We further reduce the denoising cost through progressive distillation, following prior work on guided diffusion distillation. Rather than directly distilling the model to a few sampling steps, which can degrade fine details, we progressively halve the sampling budget from 32 to
1 step. At each stage, an $M$-step student is initialized from a frozen $2M$-step teacher and trained to reproduce two teacher updates in a single step, under identical conditions and ground-truth video histories. This provides a high-fidelity few-step initialization for subsequent autoregressive adaptation.

\subsection{Autoregressive Rollout Distillation}
\label{subsec:ar_distillation}

While few-step distillation enables efficient rendering, autoregressive generation remains susceptible to accumulated errors, as each window depends on previously generated frames. Existing forcing approaches~\cite{huang2026self,hu2026longlive,zhu2026causal,zhao2026causalplus} address this issue through on-policy rollout distillation, while long-context methods further model extended histories to preserve temporal consistency. However, unlike open-ended video generation, our driving NVS task benefits from persistent geometric and reference conditions derived from the recorded scene. These scene-specific anchors reduce the reliance on generated histories for maintaining scene consistency, allowing us to adopt a simpler and potentially more efficient rollout distillation strategy.

\noindent\textbf{Geometry-Anchored Rollout Distillation.}
We first generate an extended video autoregressively, where each generated window is fed back as history for subsequent generation, exposing the student to its own accumulated errors.
We then randomly sample $S=3$ consecutive windows from the rollout and concatenate their latents into a single video segment. While the student generates these windows sequentially, a frozen bidirectional teacher jointly processes the entire segment under the corresponding geometric and
reference conditions. Following the distribution matching distillation
(DMD), we compare the teacher's predictions with those of an auxiliary denoiser trained on student outputs, and use their difference to update the autoregressive student.
Since these conditions provide persistent scene information independent of the generated history, the teacher can provide scene-grounded supervision without processing the entire autoregressive rollout. This allows us to restrict distillation to short sampled segments while exposing the student to accumulated errors at different rollout stages, encouraging long-horizon consistency with limited teacher supervision cost.

\section{Experiments}
\subsection{Experiment Settings}
\begin{table*}[htbp]
\centering
\scriptsize
\setlength{\tabcolsep}{2.5pt}
\caption{Performance comparisons on the WOD dataset for short and long sequences
under lateral shifts of $\pm 1$~m and $\pm 3$~m. IQ denotes Imaging Quality, while NTA and NTL denote NTA-IoU and NTL-IoU.
The \colorbox{bestpurple}{\textbf{best}} and \colorbox{secondpurple}{second-best} results are highlighted among all compared methods within each sequence setting.}
\label{tab:performance}

\begin{tabularx}{0.85\textwidth}{cl*{12}{Y}}
\toprule
\multirow{3}{*}{\textbf{Sequence}}
& \multirow{3}{*}{\textbf{Method}}
& \multicolumn{6}{c}{\textbf{Lateral shift $\pm1$ m}}
& \multicolumn{6}{c}{\textbf{Lateral shift $\pm3$ m}} \\
\cmidrule(lr){3-8}\cmidrule(lr){9-14}

&&
\multicolumn{3}{c}{\textbf{Visual Quality}}
& \multicolumn{3}{c}{\textbf{Consistency}}
& \multicolumn{3}{c}{\textbf{Visual Quality}}
& \multicolumn{3}{c}{\textbf{Consistency}} \\
\cmidrule(lr){3-5}\cmidrule(lr){6-8}
\cmidrule(lr){9-11}\cmidrule(lr){12-14}

&&
\metrichead{FID$\downarrow$}
& \metrichead{FVD$\downarrow$}
& \metrichead{IQ$\uparrow$}
& \metrichead{NTA$\uparrow$}
& \metrichead{NTL$\uparrow$}
& \metrichead{CLIP-V$\uparrow$}
& \metrichead{FID$\downarrow$}
& \metrichead{FVD$\downarrow$}
& \metrichead{IQ$\uparrow$}
& \metrichead{NTA$\uparrow$}
& \metrichead{NTL$\uparrow$}
& \metrichead{CLIP-V$\uparrow$} \\
\midrule

\multirow{10}{*}{\shortstack{\textbf{Short}\\ (50 frames)}}
& \multicolumn{13}{>{\columncolor{gray!20}}l}{\textit{Optimization-based}} \\

& EmerNeRF~\cite{yang2024emernerf}
& 33.19 & 283.63 & 52.94 & 0.442 & 51.09 & 94.07
& 48.64 & 598.81 & 49.78 & 0.348 & 50.43 & 91.33 \\

& Street Gaussians\cite{yan2024streetgaussian}
& 42.48 & 561.24 & 63.03 & 0.537 & 49.86 & 91.23
& 92.38 & 1115.65 & 60.95 & 0.483 & 48.35 & 85.13 \\

& \multicolumn{13}{>{\columncolor{gray!20}}l}{\textit{Diffusion-based}} \\

& FreeVS~\cite{wang2025freevs}
& 31.72 & 321.50 & 61.58 & 0.537 & 52.62 & 89.22
& 41.29 & 556.97 & 61.97 & 0.522 & 52.62 & 86.96 \\

& StreetCrafter~\cite{yan2025streetcrafter}
& 30.00 & 273.87 & 59.90 & 0.557 & 52.26 & 92.31
& 41.64 & 522.34 & 60.24 & 0.501 & 52.39 & 90.02 \\

& Difix3D+~\cite{wu2025difix3d+}
& 29.63 & 314.67 & \second{70.10} & \second{0.560}
& \best{52.92} & 91.31
& 41.05 & 542.90 & \second{69.70} & \second{0.525}
& \second{52.71} & 89.39 \\

& GEN3C~\cite{ren2025gen3c}
& 46.55 & 488.37 & 55.39 & 0.380 & 50.71 & 90.93
& 59.47 & 683.10 & 54.93 & 0.276 & 49.46 & 89.21 \\

& ReCamDriving~\cite{li2025recamdriving}
& \second{26.62} & \second{245.95} & 65.94 & 0.548 & 52.54
& \second{94.34}
& \second{37.01} & \best{374.28} & 66.04 & 0.524 & 51.30
& \second{92.01} \\

\cmidrule{2-14}
& \textbf{Ours}
& \best{23.85} & \best{230.74} & \best{70.25}
& \best{0.562} & \second{52.88} & \best{94.73}
& \best{35.84} & \second{466.40} & \best{70.27}
& \best{0.526} & \best{52.84} & \best{92.58} \\

\midrule
\multirow{9}{*}{\shortstack{\textbf{Long}\\(152 frames)}}
& \multicolumn{13}{>{\columncolor{gray!20}}l}{\textit{Optimization-based}} \\

& EmerNeRF~\cite{yang2024emernerf}
& 20.38 & 178.23 & 52.92 & 0.499 & 49.82 & \second{94.41}
& 30.07 & 412.38 & 50.11 & 0.424 & 49.42 & \second{91.97} \\

& Street Gaussians\cite{yan2024streetgaussian}
&23.87 & 351.25 & 62.27 & 0.557 & 49.06 & 92.19
& 61.73 & 827.07 & 60.33 & 0.503 & 47.93 & 87.15 \\

& \multicolumn{13}{>{\columncolor{gray!20}}l}{\textit{Diffusion-based}} \\

& FreeVS~\cite{wang2025freevs}
& 19.23 & 161.55 & 60.60 & 0.577 & 51.07 & 89.05
& 24.85 & 274.22 & 61.25 & 0.561 & 51.16 & 86.81 \\

& StreetCrafter~\cite{yan2025streetcrafter}
& 18.25 & \second{143.09} & 59.54 & \second{0.593} & 50.86 & 92.43
& 24.85 & 264.11 & 59.80 & 0.561 & 50.95 & 90.20 \\

& Difix3D+~\cite{wu2025difix3d+}
& \second{17.84} & 156.29 & \second{69.08} & 0.592
& \second{51.23} & 91.66
& \second{24.22} & \second{259.13} & \second{69.62} & \best{0.571}
& \best{51.40} & 89.79 \\

& GEN3C~\cite{ren2025gen3c}
& 34.96 & 409.12 & 56.41 & 0.438 & 49.81 & 90.73
& 44.90 & 537.61 & 55.74 & 0.307 & 49.00 & 88.86 \\

\cmidrule{2-14}
& \textbf{Ours}
& \best{13.56} & \best{129.08} & \best{69.69}
& \best{0.594} & \best{51.25} & \best{94.76}
& \best{20.17} & \best{227.25} & \best{69.70}
& \second{0.563} & \second{51.32} & \best{92.78} \\

\bottomrule
\end{tabularx}
\end{table*}

\begin{table}[t]
    \centering
    \caption{Efficiency comparison on WOD. Backbone sizes are reported in billions of parameters (B), and NFE denotes the number of denoising network evaluations per generation window, counting classifier-free guidance branches. All runtimes are measured for 121 frames (12s) generation on an NVIDIA H20 GPU, excluding VAE encoding and decoding.}
    \label{tab:efficiency}
    \resizebox{0.9\columnwidth}{!}{%
    \begin{tabular}{@{}lccccc@{}}
        \toprule
        Method & Params. (B) & NFE
        & Time $\downarrow$
        & FID$\downarrow$ & FVD$\downarrow$ \\
        \midrule

        FreeVS~\cite{wang2025freevs}
        & 1.5 & 50 & 188.42 s & 56.12 & 168.74 \\

        StreetCrafter~\cite{yan2025streetcrafter}
        & 1.5 & 100 & 824.61 s & 24.57 & 153.34 \\

        GEN3C~\cite{ren2025gen3c}
        & 7.0 & 70 & 1122 s & 36.72 & 434.47 \\

        ReCamDriving~\cite{li2025recamdriving}
        & 1.3 & 50 & 1576.58 s & 18.64 & 164.39 \\

        \midrule

        & & 1 & 14.35 s & 25.16 & 222.01 \\

        Ours-1.3B & 1.3
        & 2 & 25.00 s & 16.84 & 159.86 \\

        & & 4 & 46.74 s & 15.78 & 147.46 \\

        \midrule

        & & 1 & 70.71 s & 23.72 & 219.28 \\

        Ours-14B & 14
        & 2 & 137.37 s & 18.61 & 183.70 \\

        & & 4 & 269.94 s & 14.88 & 135.89 \\
        \bottomrule
    \end{tabular}%
    }
\end{table}

\begin{figure*}[t]
    \vspace{-10pt}
    \centering
    \includegraphics[width=0.95\textwidth]{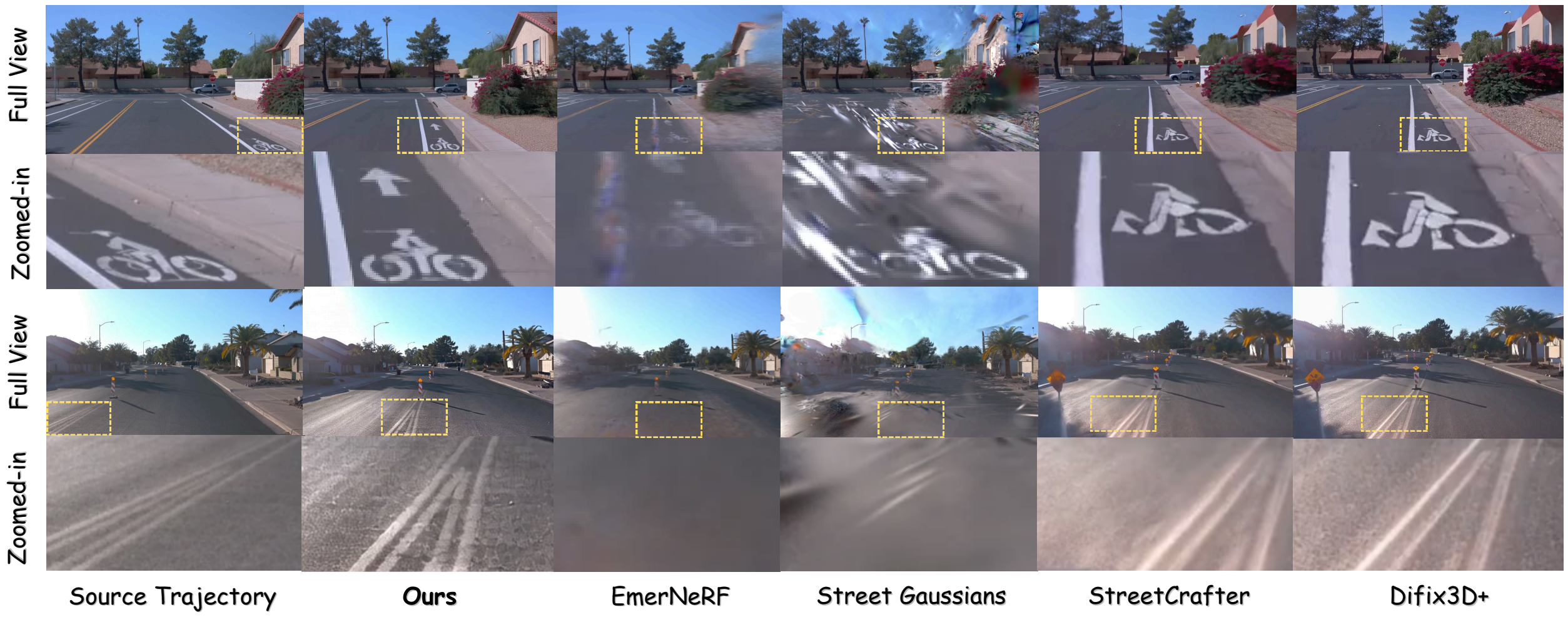}
    \caption{Qualitative comparisons on WOD~\cite{waymo_dataset} along novel trajectories against optimization-based methods (NeRF and 3DGS) and diffusion-based methods. Zoomed-in views of the highlighted regions reveal differences in geometric consistency and visual detail.}
    \label{fig:baseline_comparison}
     \vspace{-10pt}
\end{figure*}



\noindent\textbf{Implementation Details.}
We build \methodname{} upon the pretrained Wan2.1-T2V-1.3B and Wan2.1-T2V-14B models~\cite{wan2025wan}. During training, we sample three consecutive autoregressive windows, each comprising five video latent frames, conditioned on three history latent frames and two reference images. We first train the models at 384p resolution for 20k steps, followed by fine-tuning at 480p resolution for an additional 10k steps. Further details are provided in the supplementary material.

\subsection{Novel View Synthesis Quality}
\noindent\textbf{Baselines.}
We compare \methodname{} with seven representative baselines, 
categorized into two groups based on their technical design:
(1) Optimization-based methods:
EmerNeRF~\cite{yang2024emernerf} models dynamic driving scenes using decoupled static-dynamic neural fields with scene flow, while Street Gaussians~\cite{yan2024streetgaussian} represents static backgrounds and dynamic actors using 3D Gaussian splatting. (2) Diffusion-based methods:
Difix3D+~\cite{wu2025difix3d+} uses a single-step diffusion model to restore coarse renderings.
FreeVS~\cite{wang2025freevs} and StreetCrafter~\cite{yan2025streetcrafter} condition video diffusion on colored LiDAR point projections;
GEN3C~\cite{ren2025gen3c} leverages renderings from a depth-derived 3D cache;
and ReCamDriving~\cite{li2025recamdriving} combines source-video conditioning with camera trajectories and target 3DGS renderings.

\noindent\textbf{Datasets and Metrics.}
We evaluate on 20 diverse scenes from the  Waymo Open Dataset (WOD)~\cite{waymo_dataset} test split, comprehensively covering subtle (1m) to large trajectory shifts (3m) across both short-term (nearly 5s) and long-horizon rollouts (nearly 15s) and evaluate generation performance across two key dimensions: 
(1) Visual Quality, using Fr\'echet Inception Distance (FID)~\cite{heusel2017fid}, Fr\'echet Video Distance (FVD)~\cite{unterthiner2018fvd}, and Imaging Quality (IQ)~\cite{huang2024vbench} to evaluate image- and video-level distributional fidelity; and 
(2) Consistency, following DriveDreamer4D~\cite{zhao2025drivedreamer4d} to compute NTA-IoU and NTL-IoU for target-view agent and lane geometric alignment, alongside frame-wise CLIP similarity (CLIP-V)~\cite{li2025recamdriving} to measure cross-view semantic consistency as proposed in ReCamDriving~\cite{li2025recamdriving}.

\noindent\textbf{Quantitative Comparisons.} 
\Cref{tab:performance} reports quantitative results on WOD across short and long sequences under $\pm 1\,\mathrm{m}$ and $\pm 3\,\mathrm{m}$\ m shifts. Long-sequence results for ReCamDriving are omitted because its fixed temporal positional embeddings prevent generation beyond its supported temporal window. Our method outperforms the baselines on most metrics. Although ReCamDriving achieves lower FVD on short $\pm3$ m sequences, it incurs a substantial computational overhead and struggles to scale to long-horizon scenarios. Similarly, while Difix3D+ yields slightly higher NTL-IoU via frame-wise refinement, our joint multi-frame modeling achieves superior overall visual quality and temporal consistency.
\Cref{tab:efficiency} further evaluates the quality--efficiency trade-off on the same WOD dataset, with all methods evaluated on 121-frame generation. With just two denoising steps per window, our 1.3B variant outperforms FreeVS in both FID and FVD, and achieves much better FID and comparable FVD to StreetCrafter at about $33\times$ faster speed. With four steps, it surpasses all baselines on both metrics while remaining substantially faster. Even with a single denoising step, our model outperforms GEN3C in both FID and FVD at substantially lower runtime, while the four-step 14B variant achieves the best FID and FVD among the compared configurations.

\noindent\textbf{Qualitative Comparisons.}
As illustrated in \cref{fig:baseline_comparison}, we present qualitative comparisons along novel trajectories laterally shifted by 3 m from the source trajectories. We first compare our method with representative optimization-based methods, including EmerNeRF and Street Gaussians. These methods suffer from degraded rendering quality when extrapolating to laterally shifted viewpoints, where insufficient multi-view observations lead to blurred road markings and distorted textures. We also compare against diffusion-based methods, which can produce unclear lane markings and inconsistent road structures, particularly in the second example. In contrast, our method produces clearer road markings and more coherent lane boundaries across both examples, better preserving fine details and the overall road layout under lateral viewpoint shifts. More comprehensive visual comparisons with all the baselines and results on longer trajectories are provided in the supplementary material.


\subsection{Long-Horizon Comparison}
\label{sec:long-horizon_comparison}
\noindent\textbf{Baselines.} Existing diffusion-based driving NVS methods do not explicitly target long-horizon generation. We therefore select HorizonDrive~\cite{zhang2026horizondrive} and Epona~\cite{zhang2025epona}, two closely related methods for long-horizon driving video generation, as baselines.
HorizonDrive introduces scheduled rollout recovery and teacher rollout distribution matching distillation (DMD) to mitigate error accumulation during autoregressive generation. Epona is an autoregressive diffusion world model that decouples temporal dynamics modeling from visual generation and integrates trajectory prediction with video synthesis.

\noindent\textbf{Datasets and Metrics.}
We evaluate 55-second rollouts on 15 independent front-view sequences from our internal dataset, using valid trajectories that shift laterally by 2\,m and then return: leftward and rightward.
All generative models use pretrained checkpoints, with no training or fine-tuning on this internal dataset. We report FID~\cite{heusel2017fid} and FVD~\cite{unterthiner2018fvd} over the generated sequences, excluding the initial 10 frames.

\begin{table}[t]
\centering
\caption{Full-55-second performance under novel trajectories.}
\label{tab:novel_fid_fvd}
\resizebox{0.8\columnwidth}{!}{%
\begin{tabular}{lcccc}
\toprule
& \multicolumn{2}{c}{Left 2m return}
& \multicolumn{2}{c}{Right 2m return} \\
\cmidrule(lr){2-3}\cmidrule(lr){4-5}
Method & FID $\downarrow$ & FVD $\downarrow$
       & FID $\downarrow$ & FVD $\downarrow$ \\
\midrule
Epona        & 214.97 & 2267.75 & 170.93 & 1392.92 \\
HorizonDrive & 92.29  & 616.03  & 100.33 & 545.58  \\
Ours (14B)   & \best{23.08} & \best{112.74}
             & \best{24.67} & \best{125.97} \\
\bottomrule
\end{tabular}%
}
\smallskip
\end{table}


\begin{figure}[t]
    \centering
    \includegraphics[width=\columnwidth]{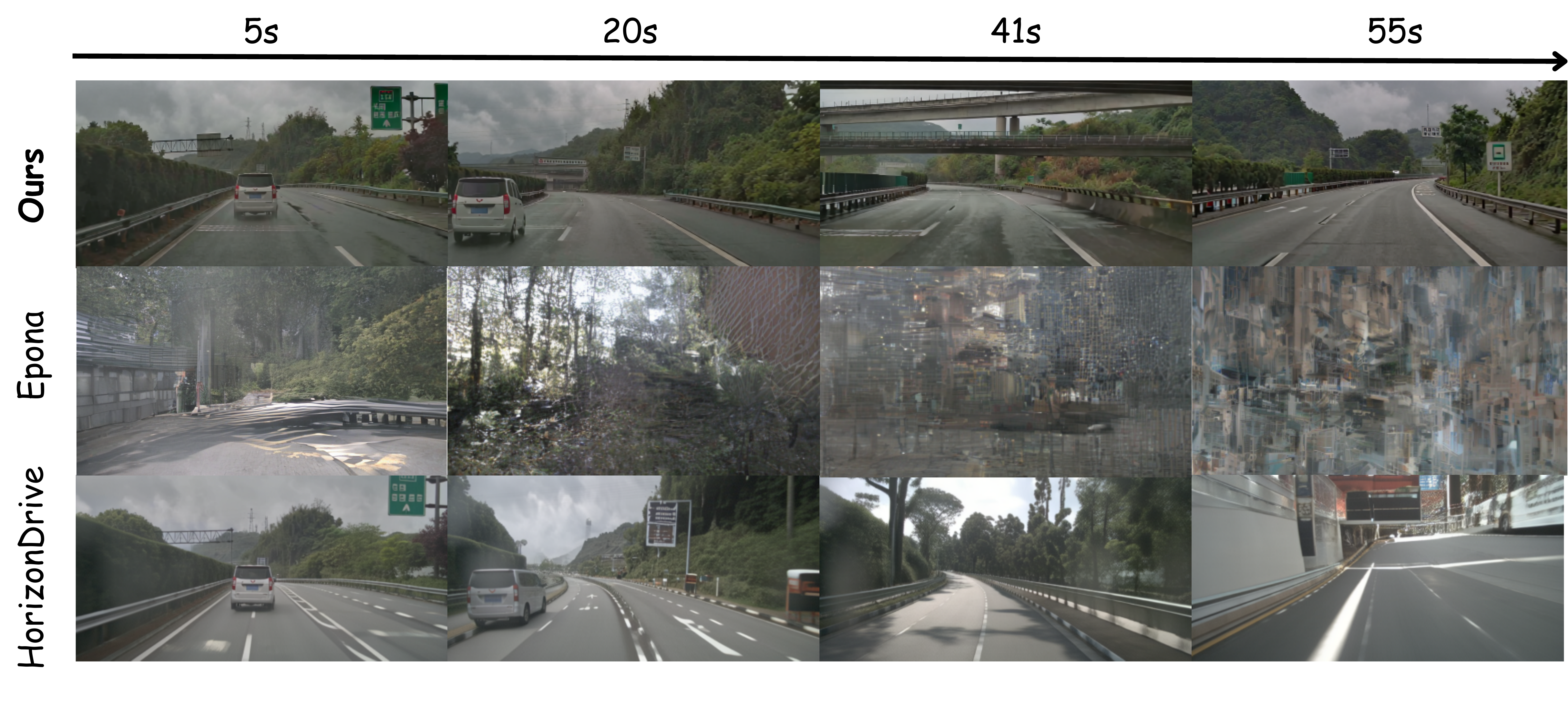}
    \caption{Qualitative comparison over a long-horizon rollout at 5, 20, 41, and 55 seconds. DriveRender preserves scene structure and visual details throughout the sequence, while other methods exhibit visual artifacts over time.}
    \label{fig:long_horizon}
     \vspace{-10pt}
\end{figure}

\noindent\textbf{Quantitative and Qualitative Comparisons.} 
As shown in ~\cref{tab:novel_fid_fvd}, our method achieves substantially lower FID and FVD than Epona and HorizonDrive under both leftward and rightward shift-and-return trajectories. These results indicate better visual quality and video-level fidelity over the full 55-second rollouts under novel camera trajectories.
The qualitative results in \cref{fig:long_horizon} further show better preservation of scene structure and visual details over time. Together, these results demonstrate stable long-horizon generation on an unseen dataset and suggest that explicit geometric conditioning helps constrain structural drift by providing a persistent reference throughout the rollout.

\begin{figure*}[htbp]
    \centering
    \includegraphics[width=\textwidth]{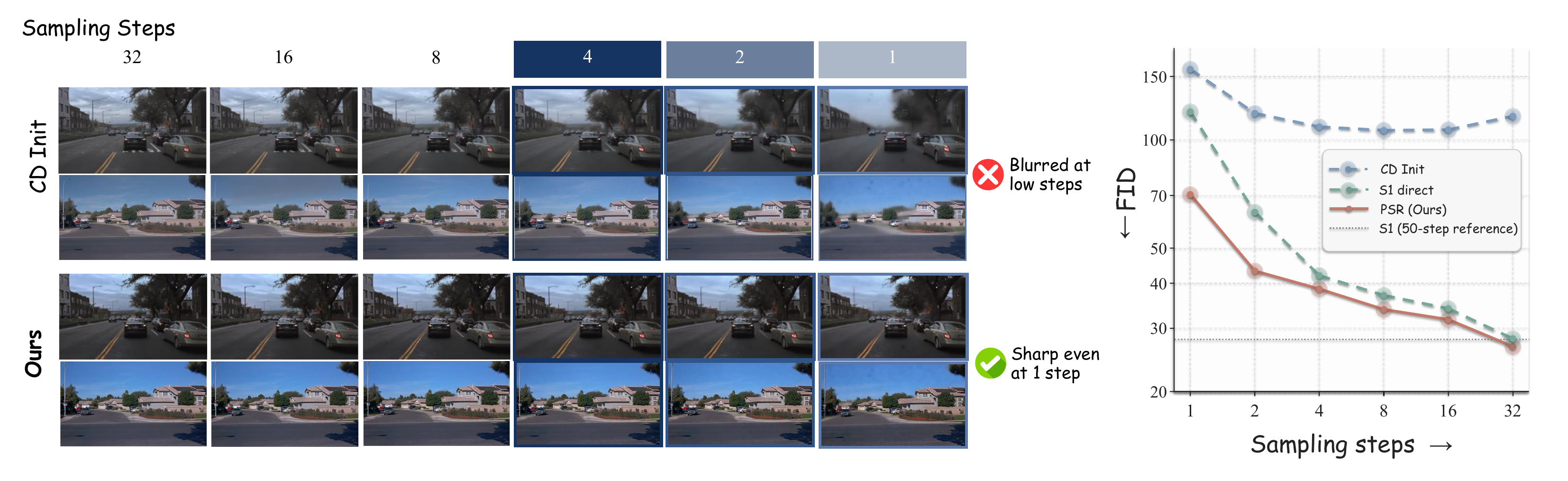}
    \caption{Comparisons on progressive step reduction.
        \textbf{Left:} CD Init produces blurry results at low sampling
        budgets, while PSR preserves sharper visual details even
        at a single step.
      \textbf{Right:} FID curves across sampling budgets. CD Init denotes initialization via consistency distillation; S1 direct denotes direct sampling from the checkpoint obtained after joint text and geometry guidance distillation, without progressive step reduction; S1 (50-step reference) denotes sampling from the same checkpoint with 50 denoising steps.
    }
    \label{fig:s2_comparison}
     \vspace{-10pt}
\end{figure*}

\begin{figure}[htbp]
    \centering
    \includegraphics[width=0.95\columnwidth]{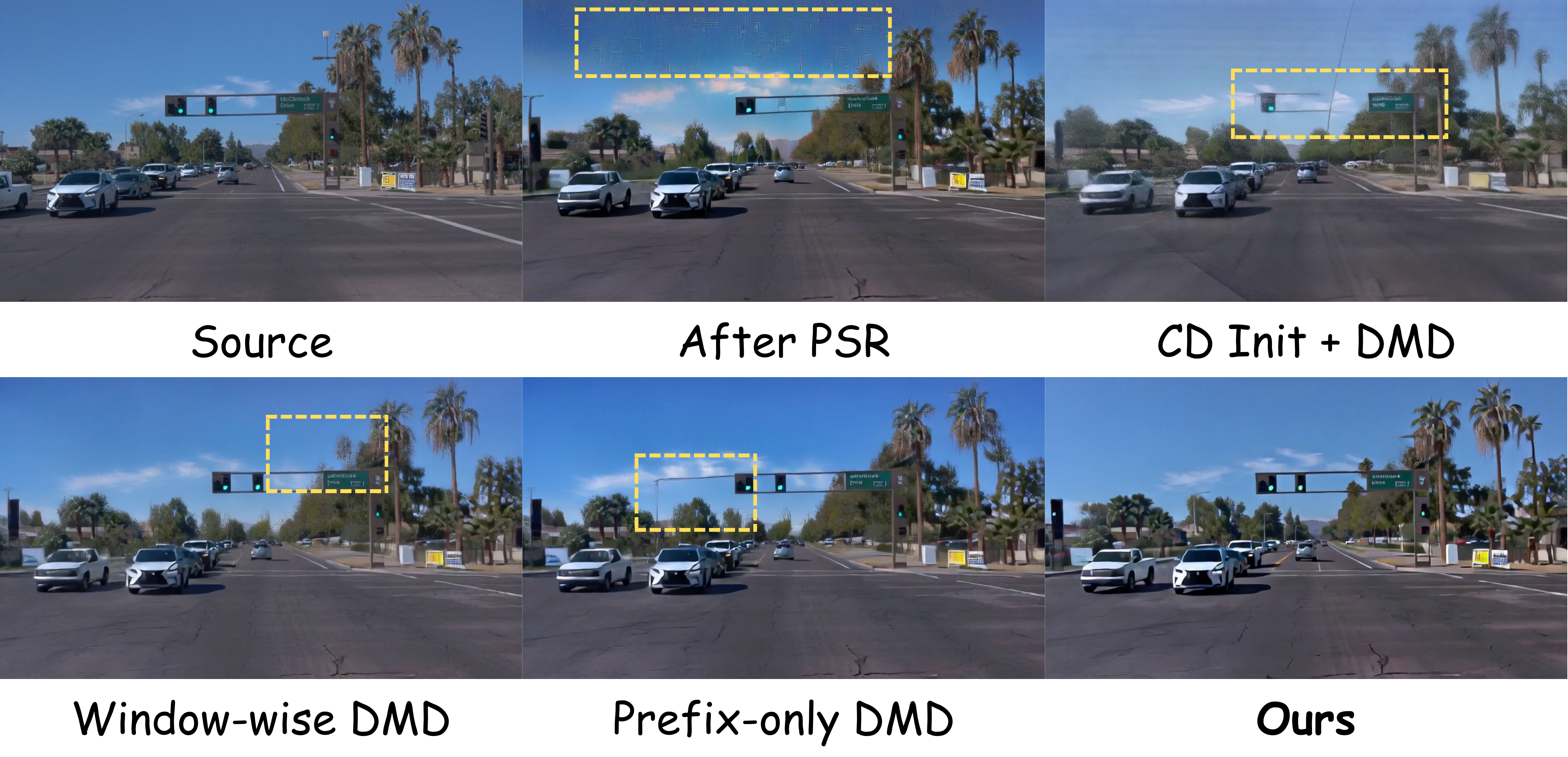}
    \caption{Comparison of distillation strategies on WOD.}
    \label{fig:s3_comparison}
    \vspace{-10pt}
\end{figure}

\subsection{Ablation Study}
\noindent\textbf{Ablation on Reference Conditioning.} \cref{tab:reference_ablation} shows that reference image conditioning consistently improves visual quality and temporal consistency, suggesting that reference frames provide complementary appearance and scene information that benefits scene-consistent rendering.


\begin{table}[htbp]
    \centering
    \caption{Ablation of reference image conditioning.}
    \label{tab:reference_ablation}
    \resizebox{0.9\linewidth}{!}{%
        \begin{tabular}{lcccc}
            \toprule
            Reference
            & FID $\downarrow$ & FVD $\downarrow$
            & CLIP-V $\uparrow$ & NTL $\uparrow$ \\
            \midrule
            Disabled
            & 68.32 & 669.03 & 83.59 & 63.72 \\
            Enabled \textbf{(Ours)}
            & \best{32.23} & \best{377.57}
            & \best{91.23} & \best{73.75} \\
            \bottomrule
        \end{tabular}%
    }
\end{table}

\noindent\textbf{Ablation on Distillation Strategy.} 
\begin{table}[t]
    \centering
    \caption{
        Ablation of distillation strategies on 17 challenging
        WOD scenes with randomly sampled lateral shifts of
        3--4\,m, evaluated over the first 114 and 190 frames.
        Prefix sampling uses only the initial three-window rollout;
        extended sampling includes segments beyond this prefix.
        Joint supervision evaluates three consecutive windows together.
        NTA denotes NTA-IoU.
    }
    \label{tab:alc_ablation}
    \scriptsize
    \setlength{\tabcolsep}{3pt}
    \renewcommand{\arraystretch}{1.20}

    \resizebox{\columnwidth}{!}{%
    \begin{tabular}{@{}llll*{6}{c}@{}}
        \toprule
        & & & &
        \multicolumn{3}{c}{114 frames} &
        \multicolumn{3}{c}{190 frames} \\
        \cmidrule(lr){5-7}
        \cmidrule(lr){8-10}

        Setting
        & Init.
        & \shortstack[l]{Rollout\\sampling}
        & \shortstack[l]{DMD\\supervision}
        & FVD$\downarrow$ & CLIP-V$\uparrow$ & NTA$\uparrow$
        & FVD$\downarrow$ & CLIP-V$\uparrow$ & NTA$\uparrow$ \\
        \midrule

        \rowcolor{gray!20}
        CD Init + DMD
        & CD & Extended & Joint
        & 684.49 & 83.92 & \second{0.423}
        & 523.42 & 83.80 & 0.471 \\
        \midrule

        After PSR
        & PSR & -- & --
        & 656.08 & 85.21 & 0.418
        & 528.63 & 81.45 & 0.470 \\

        Prefix-only DMD
        & PSR & Prefix & Joint
        & 619.80 & 87.66 & 0.422
        & 489.12 & 86.89 & \second{0.477} \\

        Window-wise DMD
        & PSR & Extended & Window-wise
        & \second{601.44} & \second{87.93} & 0.415
        & \second{461.13} & \second{87.01} & 0.473 \\
        \midrule

        \textbf{Ours}
        & PSR & Extended & Joint
        & \best{595.76} & \best{88.09} & \best{0.430}
        & \best{420.15} & \best{87.16} & \best{0.478} \\
        \bottomrule
    \end{tabular}%
    }
\end{table}
\cref{tab:alc_ablation} examines the effects of initialization, rollout sampling, and DMD supervision. Our full method achieves the best performance across all reported metrics at both evaluation horizons. Replacing our initialization with CD Init while retaining the same extended sampling and joint supervision degrades FVD, CLIP-V, and NTA, indicating that a stronger initialization benefits subsequent distillation. As shown in \cref{fig:s2_comparison}, PSR preserves sharper details and achieves lower FID than CD Init as sampling steps decrease, highlighting its robustness under aggressive step reduction. Starting from the same PSR initialization, Prefix-only DMD improves all metrics over After PSR, demonstrating the benefit of additional distillation, but remains inferior to our full method, highlighting the importance of sampling beyond the initial rollout prefix. Meanwhile, Window-wise DMD improves FVD and CLIP-V over prefix-only supervision but reduces NTA, indicating that improved visual and temporal scores alone do not guarantee stronger geometric consistency. Jointly supervising consecutive windows improves all three metrics over window-wise supervision, with a more pronounced FVD advantage at the longer evaluation horizon. ~\cref{fig:s3_comparison} provides complementary qualitative evidence, showing better preservation of scene structures in the highlighted regions with our full method. The full method also exhibits a smaller decline in CLIP-V than the model without DMD, consistent with reduced appearance drift over longer rollouts. Together, these results support the complementary roles of a strong initialization, extended rollout sampling, and joint cross-window supervision in maintaining long-horizon rendering quality and consistency.

\section{Conclusion}
In this paper, we presented \textbf{\methodname{}}, a geometry-grounded autoregressive video diffusion framework for faithful, long-horizon, and efficient novel-view synthesis in driving scenes. By combining coverage-aware point-cloud warping with viewpoint-aware reference video conditioning, our method anchors generation to the recorded environment, suppressing geometric hallucinations and preserving fine details under large viewpoint shifts. Furthermore, our staged distillation pipeline integrating joint guidance distillation with geometry-anchored rollout supervision enables continuous autoregressive synthesis with minimal drift in just 1--4 denoising steps. Extensive experiments confirm that \methodname{} outperforms prior reconstruction and generative baselines in visual fidelity, temporal coherence, and inference efficiency, offering a scalable solution for autonomous driving simulation.


\section{LLM Usage Statement}
The authors used ChatGPT to assist with English language editing and grammatical corrections in the manuscript. All AI-assisted revisions were reviewed and verified by the authors, who take full responsibility for the final content.
{
    \small
    \bibliographystyle{ieeenat_fullname}
    \bibliography{main}
}

\clearpage
\setcounter{page}{1}
\maketitlesupplementary
\appendix
\renewcommand{\figurename}{Fig.}
\renewcommand{\thefigure}{S\arabic{figure}}
\setcounter{figure}{0}

\renewcommand{\tablename}{Table}
\renewcommand{\thetable}{S\arabic{table}}
\setcounter{table}{0}

\section{Overview}
In this supplementary material, we provide comprehensive details regarding the implementation, training, and evaluation of our proposed method. The document is organized as follows:

\begin{itemize}
    \item \Cref{sec:method} elaborates on the detailed model architecture and the distillation procedure.
    \item \Cref{sec:training_details} details the data curation pipeline and training configurations.
    \item \Cref{sec:additional_baseline_comparisons} reports additional quantitative comparisons against state-of-the-art baselines to demonstrate the effectiveness of our approach.
    \item \Cref{sec:additional_ablations} presents additional ablation studies to further validate our key design choices.
    \item \Cref{sec:additional_visualizations} showcases extensive qualitative visualizations across diverse driving scenarios.
\end{itemize}

\section{Method}
\label{sec:method}
\subsection{Model Architecture}
\label{sec:architecture}
Our model consists of a frozen causal video VAE, a text encoder,
and a pretrained video diffusion transformer
(DiT)~\cite{peebles2023scalable,wan2025wan}.
For generation window $k$, the DiT receives four latent streams:
recent history $\mathbf{h}_k$, a noisy target chunk
$\mathbf{z}_k^\tau$, projected geometry $\mathbf{g}_k$, and posed
reference images $\mathbf{r}_k$, where $\tau$ denotes the diffusion
timestep.
We patchify the four streams and concatenate their tokens:
\begin{equation}
    \mathbf{X}_k^\tau
    =
    [\mathbf{H}_k;\mathbf{Z}_k^\tau;
     \mathbf{G}_k;\mathbf{R}_k].
\end{equation}
History, target, and geometry use separate patch embeddings,
while reference images share the history embedding.
Each reference image is independently VAE-encoded into a single
latent frame.
Rendering-validity and motion masks are embedded and added to
the geometry tokens.
Joint self-attention integrates information across the four
streams, while cross-attention incorporates text conditioning.
The DiT predicts the velocity field for the target latents,
and the sampler updates only $\mathbf{z}_k^\tau$, keeping all
conditioning latents fixed throughout denoising.

\noindent\textbf{Autoregressive Generation.}
We generate video in chunks of $L_z$ latent frames, retaining
up to $L_h$ recent latent frames as history.
After denoising window $k$, we update the history as
\begin{equation}
    \mathbf{h}_{k+1}
    =
    \operatorname{Tail}_{L_h}
    \bigl([\mathbf{h}_k;\hat{\mathbf{z}}_k]\bigr),
\end{equation}
where $\hat{\mathbf{z}}_k$ is the generated clean latent chunk.
The operator $\operatorname{Tail}_{L_h}$ retains the latest
$L_h$ latent frames, or all available frames if fewer are present.
Generated latents therefore serve directly as context for
subsequent windows, without decoding and re-encoding.

For streaming execution, we maintain separate causal VAE feature
caches for geometry encoding and generated-video decoding.
These caches allow geometry frames to be encoded incrementally
and generated chunks to be decoded as they become available,
while preserving temporal continuity across window boundaries.
The causal VAE encodes the first video frame into one latent
frame and each subsequent group of four frames into another.
Accordingly, a chunk of $L_z$ latent frames decoded from the
sequence head with an empty decoder cache produces $4L_z-3$
video frames.
Subsequent full chunks, decoded using the retained cache,
produce $4L_z$ frames each.

\noindent\textbf{Frame-Aligned Positional Encoding.}
We assign structured 3D rotary positional embeddings (RoPE)~\cite{su2024roformer}
to the concatenated tokens to encode temporal and spatial
relationships across streams.
Let $H_k$ denote the number of historical latent frames
available in window $k$.
Using zero-based indices, a history token at latent frame $i$
and spatial patch $(u,v)$ receives coordinates
\begin{equation}
    \mathbf{p}^{\mathrm{hist}}_k(i,u,v)
    =
    (i,u,v).
\end{equation}
Target and geometry tokens at latent frame $j$ share coordinates:
\begin{equation}
    \mathbf{p}^{\mathrm{tgt}}_k(j,u,v)
    =
    \mathbf{p}^{\mathrm{geo}}_k(j,u,v)
    =
    (H_k+j,u,v).
\end{equation}
This assignment places the target window immediately after
its history and gives corresponding target and geometry tokens
identical positional coordinates.
Their relative RoPE offset is therefore zero, providing a
positional cue for frame- and patch-level alignment.
The two streams remain distinguishable through their input
features and separate patch embeddings.

Sparse reference images provide appearance observations from
different viewpoints and timestamps.
For reference $m$, we assign
\begin{equation}
    \mathbf{p}^{\mathrm{ref}}_k(m,u,v)
    =
    (H_k+L_z+\Delta_{\mathrm{b}}+m\Delta_{\mathrm{r}},u,v),
\end{equation}
where $\Delta_{\mathrm{b}}$ separates the reference positions
from the video context and $\Delta_{\mathrm{r}}$ spaces
individual references.
We set $\Delta_{\mathrm{b}}=8$ and $\Delta_{\mathrm{r}}=4$.
These temporal coordinates index reference observations
rather than represent their capture times.
The viewpoints of the references are encoded by their
respective camera embeddings.

\subsection{Distillation Details}
\label{supp:distillation}

We provide additional details for the few-step
distillation and autoregressive rollout distillation described
in Secs.~\ref{subsec:few-step_distillation}
and~\ref{subsec:ar_distillation}.

\noindent\textbf{Joint Text and Geometry Guidance Distillation.}
We evaluate the frozen teacher under three conditioning
configurations: full conditioning, text removed, and
geometry-related conditions removed.
The noisy target and diffusion timestep are shared across
the three evaluations, and historical context is retained.
The geometry-related conditions include projected geometry,
camera information, and reference images.

The teacher predictions are combined according to
Eq.~\ref{eq:joint_text_geometry_guidance}.
We train the fully conditioned student using the mean
squared error between its predicted velocity and this
guided teacher prediction.
The teacher remains frozen throughout training.
The resulting student incorporates both guidance signals
into a single evaluation and initializes progressive
step reduction.

\noindent\textbf{Progressive Step Reduction (PSR).}
We progressively reduce the sampling budget through
$32\rightarrow16\rightarrow8\rightarrow4\rightarrow2\rightarrow1$.
At each reduction stage, the preceding model becomes a
frozen teacher, and the student is initialized from its weights.
Teacher and student receive the same scene conditions
and ground-truth histories.

For each training example, we sample an interval
$[\tau_j,\tau_{j+1}]$ from the student's sampling schedule
and noise the ground-truth target to obtain
$\mathbf{z}_{\tau_j}$.
Starting from this latent, the teacher performs two
denoising updates over the interval and reaches
$\mathbf{z}^{\mathrm{T}}_{j,2}$.
The target velocity for the student's single update is
\begin{equation}
    \mathbf{v}^{*}_{j}
    =
    \frac{
        \mathbf{z}_{\tau_j}
        -
        \mathbf{z}^{\mathrm{T}}_{j,2}
    }{
        \sigma_{\tau_j}
        -
        \sigma_{\tau_{j+1}}
    },
\end{equation}
where $\sigma_{\tau}$ denotes the noise level.
We minimize the mean squared error between the student's
predicted velocity and $\mathbf{v}^{*}_{j}$.
This target makes one student update reproduce the net
displacement of two teacher updates.
The teacher and student use nested schedules so that their
interval endpoints coincide.
The resulting few-step model initializes autoregressive
rollout distillation.

\noindent\textbf{Geometry-Anchored Rollout Distillation.}
We sample $S=3$ consecutive windows from an autoregressive
rollout, where each window uses previously generated latents
as history.
The selected windows are concatenated into a 15-latent segment
$\widehat{\mathbf{z}}$ for joint supervision.

For each segment, we provide the frozen bidirectional teacher
with geometric projections aligned to the target views,
posed reference images, and the preceding generated history.
The projections supply spatial and appearance evidence in
the target views, while the references provide additional
appearance observations of the recorded scene.
These conditions are derived from the recorded scene
rather than previous predictions, providing scene-specific
anchors even when the generated history contains errors.
The teacher can therefore supervise the selected segment
using both scene evidence and temporal context, without
processing the entire rollout.

After adding noise to $\widehat{\mathbf{z}}$, the teacher
and an auxiliary denoiser jointly process the segment
under the same conditioning inputs.
They predict clean latents
$\widetilde{\mathbf{z}}_{\psi}$ and
$\widetilde{\mathbf{z}}_{\phi}$, respectively.
We compute the normalized DMD update direction as
\begin{equation}
    \mathbf{g}_{\mathrm{DMD}}
    =
    \frac{
        \widetilde{\mathbf{z}}_{\phi}
        -
        \widetilde{\mathbf{z}}_{\psi}
    }{
        \operatorname{mean}
        \left(
            \left|
            \widehat{\mathbf{z}}
            -
            \widetilde{\mathbf{z}}_{\psi}
            \right|
        \right)
    }.
\end{equation}
The mean is computed over latent elements within each sample.
We train the student toward the detached target
$\widehat{\mathbf{z}}-\mathbf{g}_{\mathrm{DMD}}$ using
a mean squared error surrogate.
The auxiliary denoiser is trained separately on detached
student outputs using the flow-matching objective,
while the teacher remains frozen.

Generated histories are detached between windows.
We sample one denoising exit step per rollout and share
it across windows, using its clean-latent estimate as
the window output.
Only the exit-step evaluations of the supervised windows
retain gradients.
This allows the teacher to supervise consistency across
the selected windows without backpropagating through
the full autoregressive history.


\section{Implementation Details}
\label{sec:training_details}

\noindent\textbf{Datasets Curation.} We use 798 Waymo training sequences, 700 nuScenes training sequences, and 97 PandaSet sequences, including all five cameras in Waymo and all six cameras in nuScenes and PandaSet. For each sequence, we reconstruct dense point clouds from predicted depth aligned to LiDAR measurements and transform them into a shared world coordinate system using camera calibration and poses. We aggregate static observations from the current frame and up to five preceding frames, while retaining dynamic points only from the current frame to avoid motion trails. To construct pixel-aligned training pairs, we use two complementary reprojection strategies. First, we randomly sample camera-local horizontal translations from eight candidate offsets, \(\{\pm1,\pm2,\pm3,\pm4\}\) meters, and apply them to the recorded camera trajectories. For each sampled offset, we render the reconstructed geometry from the shifted viewpoints, back-project valid rendered pixels into 3D, and reproject them into the original camera views. This double-reprojection procedure produces geometric conditioning videos with viewpoint-dependent missing regions and rendering artifacts, while the original RGB videos provide aligned supervision. Second, we construct cross-camera training pairs by reprojecting reconstructed point clouds between the three front-facing cameras, using the RGB videos from the corresponding target cameras as supervision. All videos are processed at \(480\times832\) resolution, with corresponding camera intrinsics, camera-to-world poses, rendering-validity masks, and dynamic-region masks. We additionally use Qwen3-VL-8B-Instruct to generate a short and a detailed description for each source camera sequence, randomly selecting one during training. The resulting training dataset contains 78,876 augmented video sequences.

\noindent\textbf{Training Configuration.}
We build DriveRender upon the pretrained Wan2.1-T2V-1.3B and Wan2.1-T2V-14B models~\cite{wan2025wan}. During supervised training, we sample three consecutive autoregressive windows, each comprising five video latent frames conditioned on three history latent frames and two reference images. The model is trained via teacher forcing with clean ground-truth history supplied for subsequent windows, minimizing the average flow-matching loss across all three windows. Keeping the text encoder and VAE frozen, we optimize the diffusion transformer using AdamW with a learning rate of $10^{-5}$ and a weight decay of $0.01$. Training proceeds progressively from $384 \times 672$ resolution for 20k steps to $480 \times 832$ resolution for an additional 10k steps. To enable classifier-free guidance and improve robustness to incomplete conditioning, we independently drop the text prompt, rendered point-cloud conditioning, and camera conditioning with a probability of $0.2$ each, replacing dropped prompts with empty strings, substituting dropped point-cloud latents with Gaussian noise alongside cleared masks, and zeroing out dropped camera embeddings. The two reference images are jointly omitted with a probability of $0.15$ per window. For the first window, we construct the history by replicating the initial frame and replace it with Gaussian noise with a probability of $0.8$ to facilitate generation without an initial frame anchor. Additionally, we perturb the point-cloud latents with a probability of $0.5$ using a noise interpolation coefficient of $0.1$.

Following supervised training, we employ a multi-stage distillation pipeline to accelerate inference. We first distill joint text and geometry guidance at guidance scales of $2.0$ and $1.5$ respectively, into a single-pass student for 800 steps. Next, we progressively halve the denoising budget across 32, 16, 8, and 4 steps for 1,000 steps per stage, wherein each student is initialized from the preceding checkpoint and trained to match two steps of a frozen teacher. Both joint guidance distillation and progressive distillation adopt a learning rate of $5 \times 10^{-6}$ while retaining ground-truth history conditioning and the same first-window history and reference dropout probabilities as in supervised training. Finally, we perform geometry-anchored rollout distillation for 400 steps using student-generated autoregressive histories and a frozen bidirectional teacher. Both the generator and auxiliary denoiser are optimized with AdamW at a learning rate of $2 \times 10^{-6}$, executing five auxiliary-denoiser updates per generator update after 200 warm-up steps for the auxiliary denoiser.

\section{Additional Comparisons with Baseline Methods}
\label{sec:additional_baseline_comparisons}

We evaluate camera trajectory accuracy against diffusion-based methods
on WOD dataset~\cite{waymo_dataset} and compare novel-view synthesis quality with optimization-based
methods on the synthetic XLD dataset~\cite{li2024xld}.

\subsection{Camera Trajectory Accuracy}
\label{sec:wod_camera_accuracy}
We use MegaSaM~\cite{li2025megasam} with known camera intrinsics to estimate camera poses from the generated videos. We process the 50-frame and 152-frame sequences independently. For each video, we express the estimated and target trajectories relative to their respective first poses and compute errors over the remaining frames. Rotation error (RErr) is the mean geodesic distance between corresponding rotations, measured in degrees. Translation error (TErr) is the mean Euclidean distance between corresponding camera centers, measured in meters without scale fitting. 

Our method achieves the lowest RErr and TErr among the evaluated methods at both sequence lengths and both lateral shifts (\Cref{tab:camera_accuracy}). On 152-frame sequences, it reduces RErr by 38.9\% and TErr by 12.0\% at $\pm1$\,m, and RErr by 33.3\% and TErr by 16.1\% at $\pm3$\,m, relative to the best baseline for each
metric and shift. These reductions indicate closer adherence to the prescribed relative camera motion over long sequences.

\begin{table}[t]
\centering
\caption{Camera accuracy comparison with diffusion-based methods
on WOD. Lower values are better for both RErr ($^\circ$) and TErr (m). ReCamDriving is evaluated only on short sequences because it does not support inference on 152-frame sequences.}
\label{tab:camera_accuracy}
\resizebox{\linewidth}{!}{%
\begin{tabular}{llrrrr}
\toprule
Sequence & Method
& \multicolumn{2}{c}{Lateral shift $\pm1$ m}
& \multicolumn{2}{c}{Lateral shift $\pm3$ m} \\
\cmidrule(lr){3-4}\cmidrule(lr){5-6}
& & RErr$\downarrow$ & TErr$\downarrow$
  & RErr$\downarrow$ & TErr$\downarrow$ \\
\midrule
\multirow{4}{*}{\shortstack{Short\\(50 frames)}}
& StreetCrafter & 0.550 & 9.439  & 0.566 & 9.672  \\
& Difix3D+      & 0.502 & 9.485  & 0.729 & 10.956 \\
& ReCamDriving  & 0.403 & 12.943 & 0.930 & 13.011 \\
\cmidrule(lr){2-6}
& \textbf{Ours} & \best{0.323} & \best{7.601}
                & \best{0.460} & \best{7.886} \\
\midrule
\multirow{3}{*}{\shortstack{Long\\(152 frames)}}
& StreetCrafter & 1.022 & 28.518 & 0.894 & 27.133 \\
& Difix3D+      & 0.803 & 30.103 & 0.887 & 31.269 \\
\cmidrule(lr){2-6}
& \textbf{Ours} & \best{0.491} & \best{25.109}
                & \best{0.592} & \best{22.764} \\
\bottomrule
\end{tabular}%
}
\end{table}

\begin{table*}[htbp]
    \centering
    \caption{Effect of distillation and sampling budget on the 14B model
    on WOD. NFE denotes the number of denoising network evaluations per
    generation window, counting each classifier-free guidance branch
    separately. IQ denotes image quality, and NTA and NTL denote
    NTA-IoU and NTL-IoU, respectively.}
    \label{tab:distillation_budget_14b}

    \scriptsize
    \setlength{\tabcolsep}{2.5pt}
    \renewcommand{\arraystretch}{1.12}
    \begin{tabularx}{\textwidth}{@{}clc*{12}{Y}@{}}
        \toprule
        \multirow{3}{*}{\textbf{Sequence}}
        & \multirow{3}{*}{\textbf{Model}}
        & \multirow{3}{*}{\textbf{NFE}$\downarrow$}
        & \multicolumn{6}{c}{\textbf{Lateral shift $\pm1$\,m}}
        & \multicolumn{6}{c}{\textbf{Lateral shift $\pm3$\,m}} \\
        \cmidrule(lr){4-9}\cmidrule(lr){10-15}
        & &
        & \multicolumn{3}{c}{\textbf{Visual Quality}}
        & \multicolumn{3}{c}{\textbf{Consistency}}
        & \multicolumn{3}{c}{\textbf{Visual Quality}}
        & \multicolumn{3}{c}{\textbf{Consistency}} \\
        \cmidrule(lr){4-6}\cmidrule(lr){7-9}
        \cmidrule(lr){10-12}\cmidrule(lr){13-15}
        & &
        & FID$\downarrow$ & FVD$\downarrow$ & IQ$\uparrow$
        & NTA$\uparrow$ & NTL$\uparrow$ & CLIP-V$\uparrow$
        & FID$\downarrow$ & FVD$\downarrow$ & IQ$\uparrow$
        & NTA$\uparrow$ & NTL$\uparrow$ & CLIP-V$\uparrow$ \\
        \midrule

        \rowcolor{gray!20}
        \cellcolor{white}\multirow{4}{*}{\shortstack{\textbf{Short}\\(50 frames)}}
        & Undistilled & 150
        & 19.99 & 151.51 & 69.35 & 0.564 & 52.76 & 95.09
        & 35.42 & 343.75 & 69.59 & 0.531 & 52.87 & 92.79 \\
        & Distilled & 4
        & 23.85 & 230.74 & 70.25 & 0.562 & 52.88 & 94.73
        & 35.84 & 466.40 & 70.27 & 0.526 & 52.84 & 92.58 \\
        & Distilled & 2
        & 28.42 & 309.37 & 71.08 & 0.556 & 52.89 & 94.07
        & 38.04 & 525.02 & 70.85 & 0.538 & 52.89 & 92.01 \\
        & Distilled & 1
        & 34.21 & 365.54 & 71.69 & 0.547 & 53.16 & 93.44
        & 42.40 & 545.46 & 71.37 & 0.529 & 53.17 & 91.42 \\
        \midrule

        \rowcolor{gray!20}
        \cellcolor{white}\multirow{4}{*}{\shortstack{\textbf{Long}\\(152 frames)}}
        & Undistilled & 150
        & 12.08 & 103.41 & 68.84 & 0.598 & 51.16 & 94.58
        & 21.23 & 208.00 & 68.97 & 0.566 & 51.40 & 92.24 \\
        & Distilled & 4
        & 13.56 & 129.08 & 69.69 & 0.594 & 51.25 & 94.76
        & 20.17 & 227.25 & 69.70 & 0.563 & 51.32 & 92.78 \\
        & Distilled & 2
        & 17.12 & 176.24 & 70.50 & 0.582 & 51.30 & 94.23
        & 22.83 & 263.73 & 70.29 & 0.575 & 51.34 & 92.22 \\
        & Distilled & 1
        & 22.01 & 211.90 & 71.25 & 0.587 & 51.62 & 93.77
        & 26.62 & 289.08 & 71.60 & 0.565 & 51.60 & 91.73 \\
        \bottomrule
    \end{tabularx}
\end{table*}

\subsection{Novel-View Synthesis on synthetic dataset}
\label{sec:synthetic_dataset}
We evaluate our model on the synthetic XLD dataset without fine-tuning and compare it with optimization-based NeRF and 3D Gaussian Splatting methods. We measure novel-view synthesis quality
using PSNR and LPIPS against ground-truth target-view images.As shown in \Cref{tab:xld}, our method achieves the highest PSNR and lowest LPIPS among the compared methods across all three lateral offsets. At offsets of 1, 2, and 4 meters, it improves PSNR over the strongest compared baseline at each offset by 2.07, 1.37 and 0.91\,dB, respectively. It also achieves LPIPS scores of 0.211, 0.223, and 0.265, corresponding to relative reductions of 32.8\%, 41.2\%, and 35.7\% over the lowest baseline LPIPS at each offset.
Although increasing the lateral offset degrades performance, our model maintains an advantage in both metrics.
These results demonstrate its ability to transfer to synthetic driving scenes without dataset-specific fine-tuning and synthesize faithful novel views beyond the source trajectory.

\begin{table}[htbp]
\centering
\caption{
Quantitative comparisons on the XLD dataset at lateral offsets of 1, 2, and 4 meters. We report PSNR and LPIPS against ground-truth target-view images.}
\label{tab:xld}

\resizebox{\columnwidth}{!}{%
\begin{tabular}{lcccccc}
\toprule
& \multicolumn{2}{c}{Offset-1m}
& \multicolumn{2}{c}{Offset-2m}
& \multicolumn{2}{c}{Offset-4m} \\
\cmidrule(lr){2-3}
\cmidrule(lr){4-5}
\cmidrule(lr){6-7}
Method
& PSNR$\uparrow$ & LPIPS$\downarrow$
& PSNR$\uparrow$ & LPIPS$\downarrow$
& PSNR$\uparrow$ & LPIPS$\downarrow$ \\
\midrule

\rowcolor{gray!15}
\multicolumn{7}{l}{\textbf{NeRF-based}} \\
Instant-NGP~\cite{muller2022instant}
& 23.44 & 0.346
& 22.37 & 0.386
& 20.95 & 0.443 \\
NeRFacto~\cite{tancik2023nerfstudio}
& 23.49 & 0.314
& 21.64 & 0.379
& 20.82 & 0.412 \\
\midrule

\rowcolor{gray!15}
\multicolumn{7}{l}{\textbf{Gaussian-based}} \\
3DGS~\cite{kerbl20233d}
& 23.26 & 0.334
& 22.01 & 0.396
& 19.17 & 0.460 \\
GaussianPro~\cite{cheng2024gaussianpro}
& 22.61 & 0.338
& 21.26 & 0.383
& 18.75 & 0.445 \\
\midrule

\textbf{Ours}
& \best{25.56} & \best{0.211}
& \best{23.74} & \best{0.223}
& \best{21.86} & \best{0.265} \\
\bottomrule
\end{tabular}%
}
\end{table}

\subsection{Long-Horizon Comparison}
In \cref{sec:long-horizon_comparison}, we evaluate 55-second rollouts
on 15 front-view sequences from our internal dataset using leftward
and rightward 2\,m shift-and-return trajectories. Here, we use
18-second rollouts on nuScenes to test whether DriveRender's advantage
over Epona and HorizonDrive persists across different lateral
displacements on a public dataset. We use front-camera
videos from 20 held-out scenes in the nuScenes~\cite{caesar2020nuscenes}
validation split. The evaluation comprises 46 scene--trajectory pairs
covering four maneuvers: a 1\,m or 3\,m shift to the left or right,
followed by a return to the source trajectory. All three methods
generate 18-second rollouts for the same pairs. For
Epona~\cite{zhang2025epona} and HorizonDrive~\cite{zhang2026horizondrive},
we use their released nuScenes-trained checkpoints and native
conditioning interfaces. Using recorded source-view videos as the
reference distribution, we compute FID~\cite{heusel2017fid} on
individual frames and FVD~\cite{unterthiner2018fvd} on
non-overlapping clips from the 18-second rollouts.

\begin{figure*}[htbp]
    \centering
    \includegraphics[width=\textwidth]{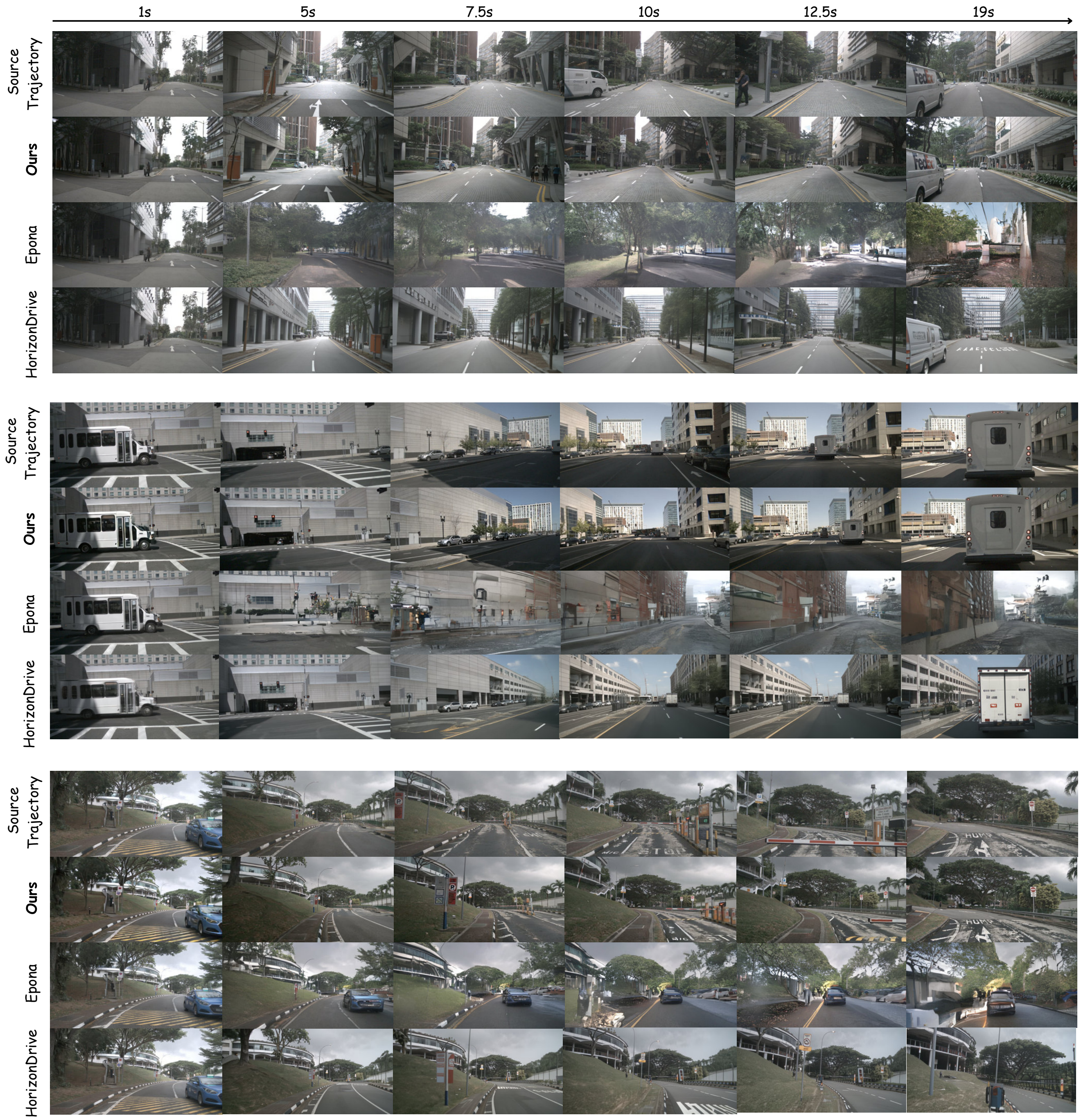}
    \caption{\textbf{Long-horizon qualitative comparison on nuScenes~\cite{caesar2020nuscenes} along a
        shift-and-return trajectory.} Columns show frames at the indicated
        timestamps. The top row shows the recorded front-camera video for
        reference, followed by DriveRender, Epona, and HorizonDrive.
        DriveRender preserves recognizable scene details across the rollout,
        including the elevated roadway and the white vehicle visible in
        the final column.}
    \label{fig:long_horizon_supp}
\end{figure*}

As shown in \cref{tab:nuscenes_novel_fid_fvd}, DriveRender achieves
lower FID and FVD than both baselines on all four shift-and-return
maneuvers. As the lateral displacement increases, its FID advantage
over HorizonDrive remains substantial, while its relative FVD
advantage narrows for both leftward and rightward shifts.
\Cref{fig:long_horizon_supp} illustrates the visual differences
late in the rollout. In the final frame, both the white vehicle and
the roadside buildings remain recognizable in DriveRender's output.
Together, these comparisons show that the gains in frame- and
clip-level fidelity observed on our internal dataset also hold on
nuScenes across the tested lateral displacements.

\begin{table}[htbp]
\centering
\caption{Performance over the 18-second generated segment on nuScenes
under novel shift-and-return trajectories.
Leftward and rightward groups contain 7 and 16 scenes, respectively,
covering 20 distinct scenes and 46 trajectory configurations in total.}
\label{tab:nuscenes_novel_fid_fvd}
\resizebox{\columnwidth}{!}{%
\begin{tabular}{lcccccccc}
\toprule
& \multicolumn{2}{c}{Left 1m return}
& \multicolumn{2}{c}{Left 3m return}
& \multicolumn{2}{c}{Right 1m return}
& \multicolumn{2}{c}{Right 3m return} \\
\cmidrule(lr){2-3}
\cmidrule(lr){4-5}
\cmidrule(lr){6-7}
\cmidrule(lr){8-9}
Method
& FID $\downarrow$ & FVD $\downarrow$
& FID $\downarrow$ & FVD $\downarrow$
& FID $\downarrow$ & FVD $\downarrow$
& FID $\downarrow$ & FVD $\downarrow$ \\
\midrule
Epona
& 93.58 & 1178.99
& 102.84 & 1418.24
& 76.88 & 927.72
& 69.26 & 885.53 \\
HorizonDrive
& 74.86 & 844.27
& 71.31 & 798.48
& 47.45 & 532.72
& 49.29 & 525.48 \\
Ours
& \best{23.48} & \best{320.26}
& \best{33.68} & \best{574.95}
& \best{18.13} & \best{292.67}
& \best{24.82} & \best{430.70} \\
\bottomrule
\end{tabular}%
}
\end{table}

\section{Additional Ablation Studies and Comparisons}
\label{sec:additional_ablations}

\paragraph{Ablation on Sampling Budget.}
\Cref{tab:distillation_budget_14b} compares the undistilled 14B model at 150 NFE with its distilled variants at 4, 2, and 1 NFE. Our distillation reduces the sampling budget of the 14B model from 150 to 4 NFE, a $37.5\times$ reduction in denoiser evaluations (\Cref{tab:distillation_budget_14b}). At $\pm3$\,m, the 4-NFE model closely matches the undistilled model's FID on 50-frame sequences (35.84 vs.\ 35.42) and improves it on 150-frame sequences (20.17 vs.\ 21.23). The 4-NFE model achieves the lowest FID and FVD among the distilled variants in all four evaluation settings. Both metrics worsen as the budget drops to 2 and 1 NFE. Even at
these lower budgets, both variants exceed the 150-NFE reference in IQ and NTL-IoU across all four settings, with the 1-NFE variant achieving the highest scores on both metrics.

\paragraph{Ablation on Joint Guidance Distillation.}
We evaluate the Stage 1 guidance formulations in \cref{tab:s1_guidance_ablation}. Standard \textit{Text CFG} retains geometric inputs but amplifies only text residuals, whereas our joint guidance explicitly incorporates a geometry-guidance direction to balance semantic priors and spatial constraints. Compared with \textit{Text CFG}, our setting $(w_t, w_g) = (2.0, 1.5)$ substantially reduces FVD by 39.8\%, 33.5\%, and 34.9\% across Waymo, nuScenes, and Argoverse 2, while improving frame-aligned CLIP-V. Among joint variants, shifting weights from $(1.0, 2.0)$ to $(2.0, 1.0)$ improves both metrics across datasets. Increasing $w_g$ to 1.5 further yields the lowest FVD on all benchmarks with minimal CLIP-V trade-offs, achieving an optimal trade-off between scene fidelity and visual quality.

\begin{table}[htbp]
    \centering
    \caption{Quantitative ablation study of joint guidance distillation on Waymo and nuScenes.
    Each pair $(w_t,w_g)$ gives the text and geometry guidance weights.}
    \label{tab:s1_guidance_ablation}
    \scriptsize
    \setlength{\tabcolsep}{2pt}
    \renewcommand{\arraystretch}{1.10}

    \resizebox{0.85\columnwidth}{!}{%
    \begin{tabular}{@{}lc*{2}{cc}@{}}
        \toprule
        \multirow{2}{*}{\textbf{Setting}}
        & \multirow{2}{*}{$(w_t,w_g)$}
        & \multicolumn{2}{c}{\textbf{Waymo}}
        & \multicolumn{2}{c}{\textbf{nuScenes}} \\
        \cmidrule(lr){3-4}\cmidrule(lr){5-6}
        & & FVD$\downarrow$ & CLIP-V$\uparrow$
          & FVD$\downarrow$ & CLIP-V$\uparrow$ \\
        \midrule
        Text CFG & $(3.0,\text{--})$
        & 612.38 & 88.02
        & 282.33 & 87.62 \\

        Joint & $(1.0,2.0)$
        & 404.95 & 90.49
        & 201.97 & 90.06 \\

        Joint & $(2.0,1.0)$
        & \second{382.98} & \best{91.34}
        & \second{194.58} & \second{90.20} \\

        \midrule
        \textbf{Ours} & $(2.0,1.5)$
        & \best{368.83} & \second{91.26}
        & \best{187.69} & \best{90.31} \\
        \bottomrule
    \end{tabular}%
    }
\end{table}

\section{Additional Visualizations}
\label{sec:additional_visualizations}

\subsection{Temporal Consistency Across Methods}
\begin{figure*}[htbp]
    \centering
    \includegraphics[width=\textwidth]{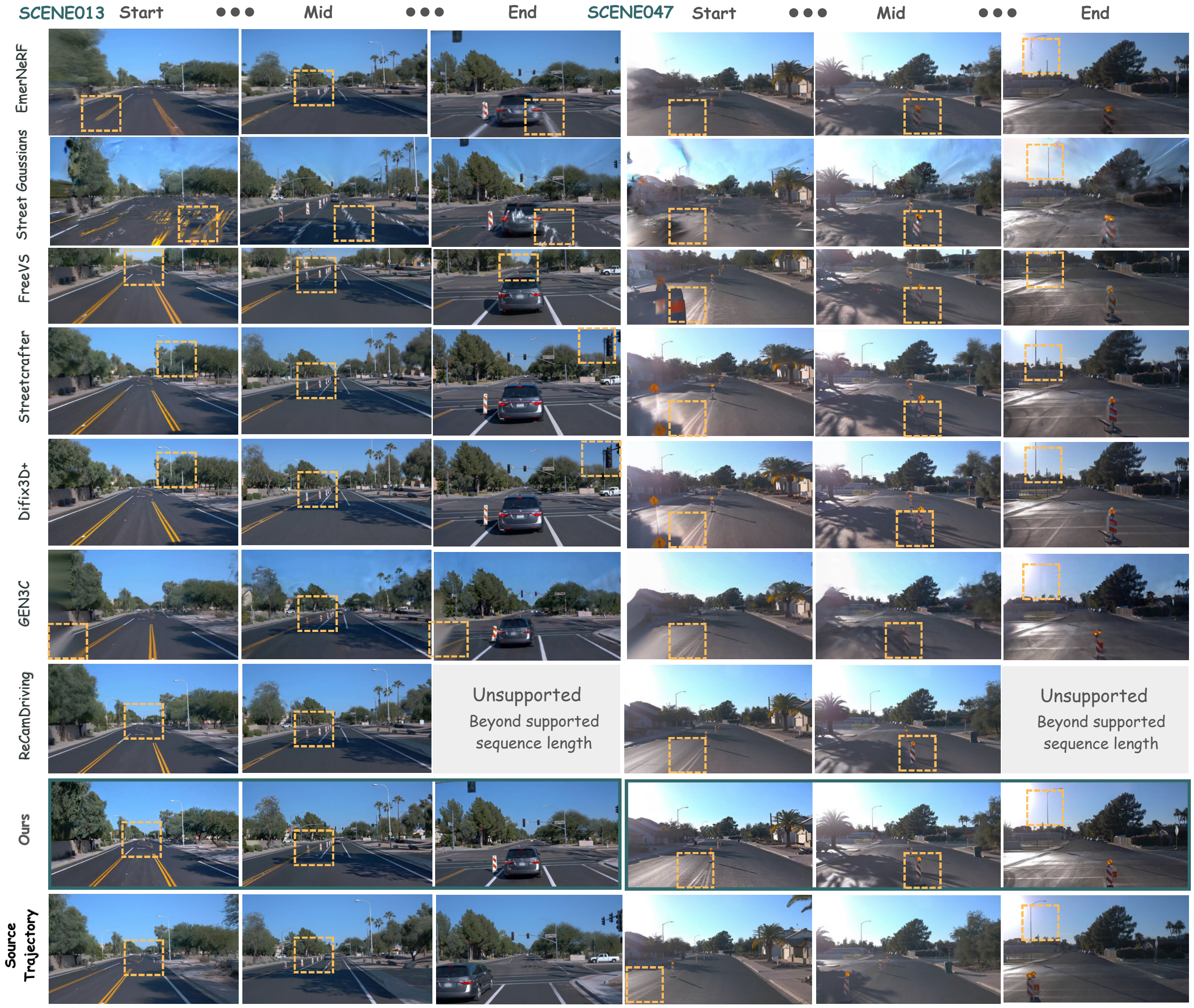}
    \caption{\textbf{Qualitative comparison with all baselines on long-horizon trajectories (up to 20\,s).}
For each scene, we show frames sampled from the start, middle, and end of the trajectory. Rows compare the results of each baseline and our method, with the corresponding source trajectory provided at the bottom for reference. Yellow dashed boxes highlight regions for close-up inspection, and ``Unsupported'' denotes frames exceeding a method's maximum supported sequence length.}
    \label{fig:nvs_supp}
\end{figure*}

To further examine temporal consistency, \Cref{fig:nvs_supp} compares our method with all baselines at the start, middle, and end of two driving trajectories lasting up to 20\,s. In \textsc{Scene013}, the lane markings
remain continuous across the sampled views as the camera approaches the intersection, and the vehicle ahead retains a clear outline in the final view. In \textsc{Scene047}, our method preserves the tree shadows on the
road as the camera moves from shade into sunlight. Several baselines introduce bright patches or streaks on the pavement in the corresponding views.

\begin{figure*}[htbp]
    \centering
    \includegraphics[width=\textwidth]{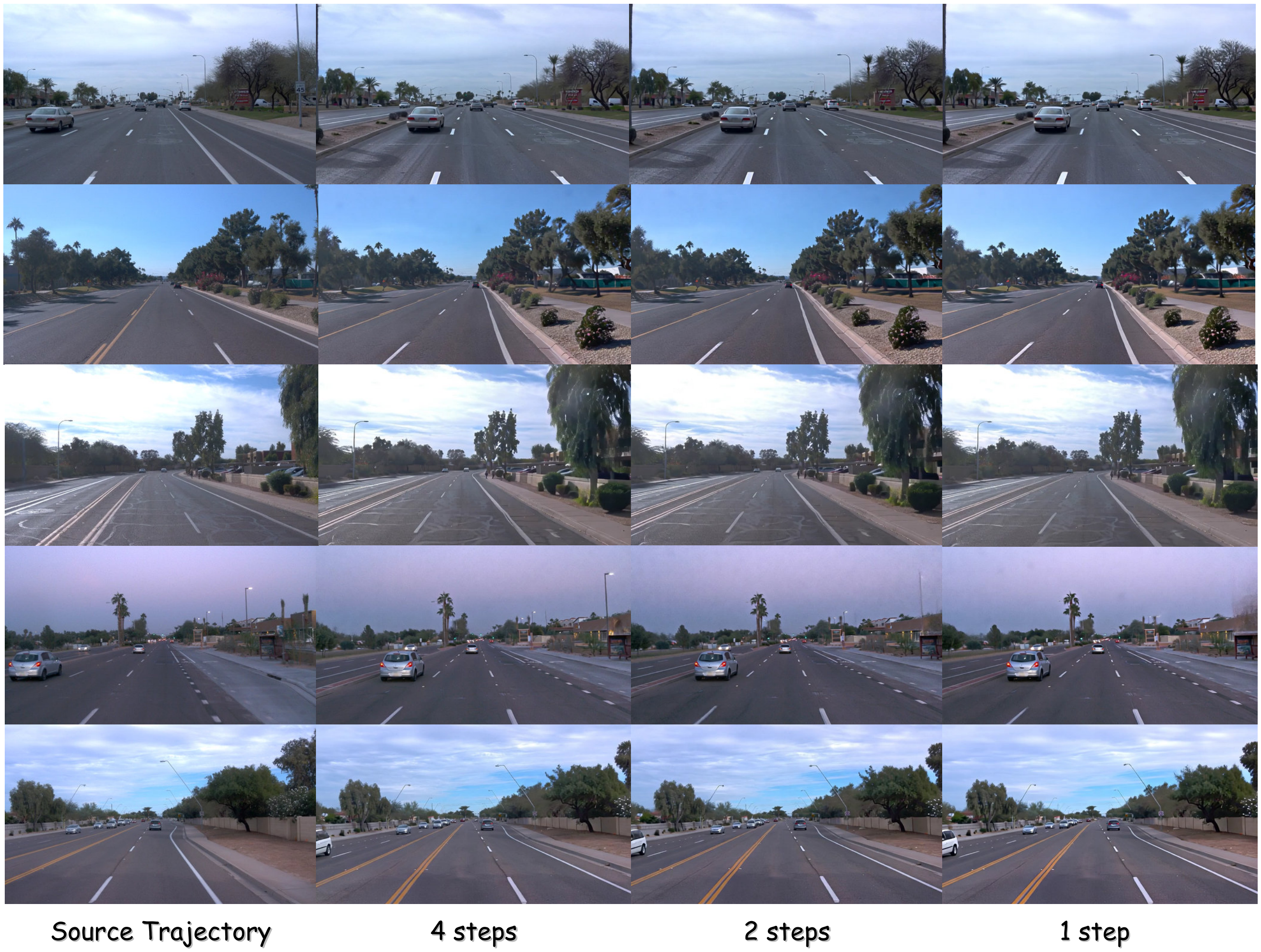}
    \caption{\textbf{Qualitative comparison across four driving scenes under different step counts with a 3\,m lateral shift}. Each row shows one scene, with columns from left to right displaying the source trajectory followed by results generated with 4, 2, and 1 denoising steps.}
    \label{fig:steps_comparison}
\end{figure*}

\subsection{High-Fidelity Rendering in 1--4 Denoising Steps}
As shown in ~\cref{fig:steps_comparison}, our method synthesizes high-fidelity novel views with as few as one denoising step. Across the four driving scenes, lane markings, vehicle outlines, and roadside trees
remain visually similar in the one-step and four-step results under the same 3\,m lateral shift.

\subsection{Long-Horizon Novel View Synthesis}
\begin{figure*}[htbp]
    \centering
    \includegraphics[width=\textwidth]{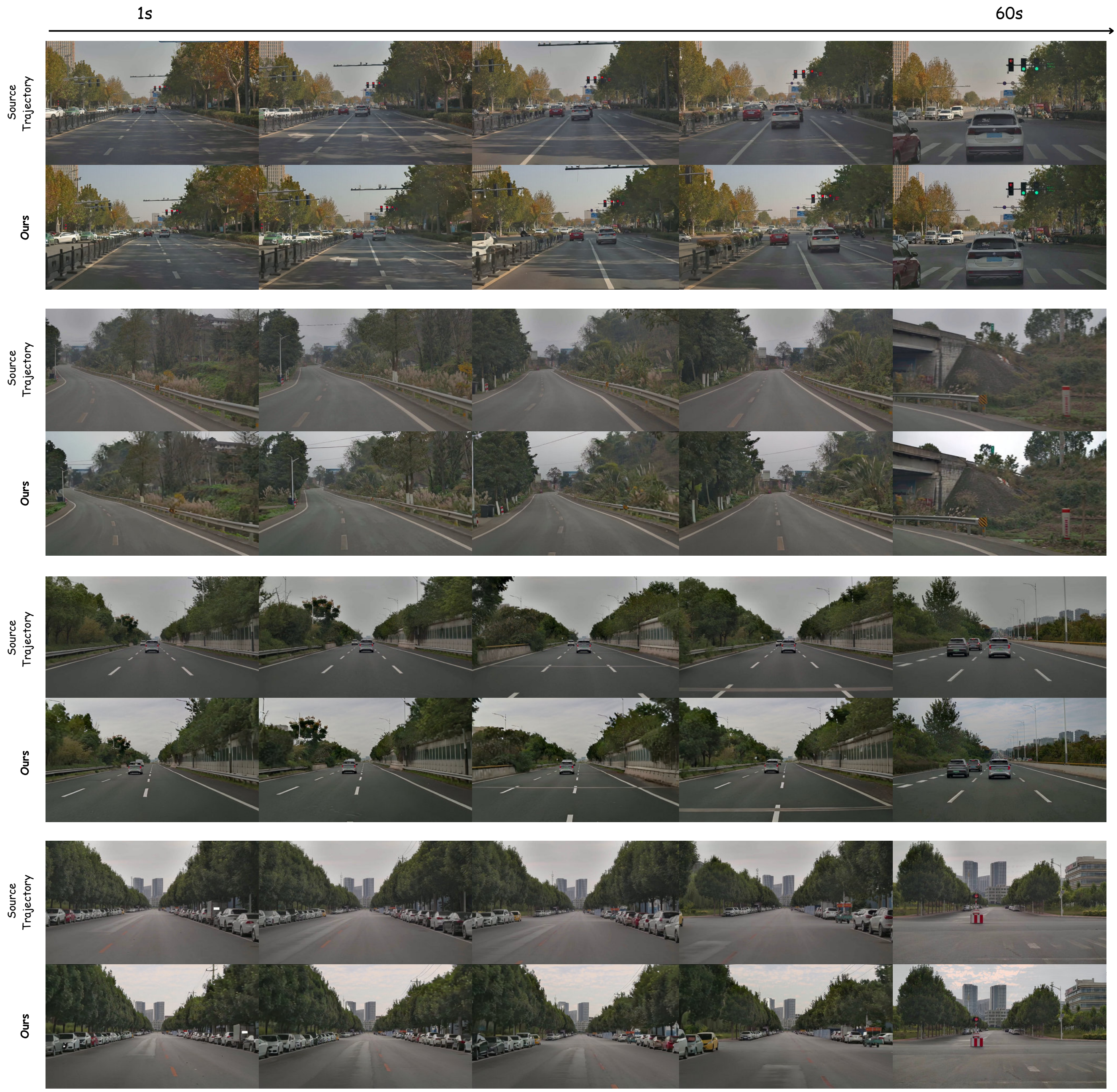}
    \caption{\textbf{Long-horizon novel view synthesis on our internal driving dataset.}
Each row pair shows frames from the source trajectory (top) and novel views
generated by our method (bottom). Columns show selected frames from
1\,s to 60\,s, ordered from left to right.}
    \label{fig:internal_dataset_supp}
\end{figure*}

We evaluate long-horizon novel view synthesis on 60-second sequences
from our internal driving dataset. ~\cref{fig:internal_dataset_supp}
shows results for four driving scenes, including an approach to a
signalized intersection, a curved road, a divided highway, and a
city street lined with parked vehicles. Across the sampled frames,
the generated views follow the changing road layout and preserve
recognizable scene features, such as the guardrail along the curve
and the rows of parked vehicles on the city street. These examples suggest that our method can preserve the spatial structure of diverse driving scenes over a 60-second horizon.

\end{document}